\documentclass[conference]{IEEEtran}
\IEEEoverridecommandlockouts

\usepackage{pgfplots}
\usetikzlibrary{external}

\usepackage{times} 
\usepackage{helvet} 
\usepackage{courier} 
\usepackage{graphicx}
\usepackage{caption} 
\usepackage{algorithm}
\usepackage{algorithmic}

\usepackage{selectp}
\usepackage{float}

\usepackage{newfloat}
\usepackage{listings}

\usetikzlibrary{patterns}

\usepackage{algorithm}
\usepackage{algorithmic}

\usepackage{selectp}
\usepackage{float}

\usepackage{newfloat}
\usepackage{listings}

\usepackage{bibentry}
\usepackage{xspace}

\usepackage{amsmath}
\DeclareMathOperator*{\argmax}{argmax}
\DeclareMathOperator*{\argmin}{argmin}
\usepackage{stmaryrd}
\usepackage{amsthm}
\usepackage{amsfonts}
\usepackage{cleveref}
\usepackage{bbold}
\usepackage{xcolor}
\usepackage{todonotes}
\usepackage{subcaption}
\usepackage{enumitem}
\usepackage{pgfplots}
\pgfplotsset{compat=1.9}
\usetikzlibrary{calc}
\usepackage{algorithm}
\usepackage{algorithmic}

\usepackage{balance}

\usepackage{subcaption}
\usepackage{cite}

\def\BibTeX{{\rm B\kern-.05em{\sc i\kern-.025em b}\kern-.08em
    T\kern-.1667em\lower.7ex\hbox{E}\kern-.125emX}}
\newtheorem{theorem}{Theorem}
\newtheorem{lemma}{Lemma}
\newtheorem{corollary}{Corollary}

\begin{document}

\title{ Learning Contextual Runtime Monitors\\
for Safe AI-Based Autonomy
\thanks{This work was partly supported by the Wallenberg AI, Autonomous Systems and Software Program funded by the Knut and Alice Wallenberg Foundation, and by the iCyPhy Center at UC Berkeley. The computations were enabled by resources provided by the National Academic Infrastructure for Supercomputing in Sweden (NAISS), partially funded by the Swedish Research Council through grant agreement no. 2022-06725. }
} 

\author{
\IEEEauthorblockN{Alejandro Luque-Cerpa}
\IEEEauthorblockA{\textit{Computer Science and Engineering} \\
\textit{Chalmers University of Technology} \\ \textit{ and University of Gothenburg}\\
Gothenburg, Sweden \\
luque@chalmers.se}
\and
\IEEEauthorblockN{Mengyuan Wang}
\IEEEauthorblockA{\textit{Computer Science and Engineering} \\
\textit{Chalmers University of Technology} \\ \textit{ and University of Gothenburg}\\
Gothenburg, Sweden \\
mengyuan@chalmers.se}
\and
\IEEEauthorblockN{Emil Carlsson}
\IEEEauthorblockA{\textit{Sleep Cycle AB}\\
Gothenburg, Sweden \\
emil.carlsson@sleepcycle.com}
\and
\IEEEauthorblockN{Sanjit A. Seshia}
\IEEEauthorblockA{\textit{Electrical Engineering and Computer Sciences} \\
\textit{University of California at Berkeley}\\
Berkeley, United States \\
sseshia@berkeley.edu}
\and
\IEEEauthorblockN{Devdatt Dubhashi}
\IEEEauthorblockA{\textit{Computer Science and Engineering} \\
\textit{Chalmers University of Technology} \\ \textit{ and University of Gothenburg}\\
Gothenburg, Sweden \\
dubhashi@chalmers.se}
\and
\IEEEauthorblockN{Hazem Torfah}
\IEEEauthorblockA{\textit{Computer Science and Engineering} \\
\textit{Chalmers University of Technology} \\ \textit{ and University of Gothenburg}\\
Gothenburg, Sweden \\
hazemto@chalmers.se}
}

\maketitle

\begin{abstract}
We introduce a novel framework for learning context-aware runtime monitors for AI-based control ensembles. Machine-learning (ML)-based controllers are increasingly deployed in (autonomous) cyber–physical systems because of their ability to solve complex decision-making tasks. However, their accuracy can degrade sharply in unfamiliar environments, creating significant safety concerns. Traditional ensemble methods aim to improve robustness by averaging or voting across multiple controllers, yet this may dilute the specialized strengths that individual controllers exhibit in different operating contexts. 
We argue that, rather than blending controller outputs, a monitoring framework should identify and exploit these contextual strengths.
In this paper, we reformulate the design of safe AI-based control ensembles as a contextual monitoring problem. A monitor continuously observes the system’s context and selects the controller best suited to the current conditions. To achieve this, we cast monitor learning as a contextual learning task and draw on techniques from contextual multi-armed bandits. Our approach comes with two key benefits: (1) theoretical safety guarantees during controller selection, and (2) improved utilization of controller diversity. We validate our framework in two simulated autonomous driving scenarios, demonstrating significant improvements in both safety and performance compared to non-contextual baselines.

\end{abstract}

\begin{IEEEkeywords}
Runtime assurance, Contextual bandits, Safe AI-based autonomy
\end{IEEEkeywords}
\maketitle

\newcommand{\Dist}{\text{Dist}}
\newcommand{\traces}[1]{\mathit{Traces(#1)}}
\newcommand{\prob}[1]{\text{Pr}(#1)}
\newcommand{\mgs}{MGS\xspace}
\newcommand{\acps}{ACPS\xspace}

\newcommand{\vcont}{V_\mathit{cont}\xspace}
\newcommand{\vsys}{V_\mathit{state}\xspace}
\newcommand{\venv}{V_\mathit{env}\xspace}
\newcommand{\vctrl}{\mathit{control}\xspace}
\newcommand{\vplant}{V_\mathit{state}\xspace}

\newtheorem{remark}{Remark}

\newcommand{\bx}{{\bf{x}}}
\newcommand{\R}{\mathbb{R}}
\newcommand{\A}{\mathcal{A}}
\newcommand{\Hess}{\mathbf{H}}
\newcommand{\btheta}{\bf{\theta}}

\section{INTRODUCTION}
Recent advancements in machine learning (ML) have been pivotal to the progress of autonomous cyber-physical systems (\acps) \cite{Bojarski2016EndTE,7778091}. ML-based models, such as neural networks, have enabled scalable solutions for complex tasks in perception, planning, and control. Growing reliance on ML-based models introduces, however, significant challenges related to safety 
\cite{amodei2016concreteproblemsaisafety,seshia-arxiv16}. ML models are inherently brittle; unanticipated changes in the \acps's environment can degrade performance and lead to faulty outcomes that may compromise system safety. This brittleness can be the consequence of various factors, related to the training algorithms used, the model architecture, and the nature of the training data. One common approach to building more robust models is through the use of ensembles: multiple models whose predictions are combined to produce a more robust output. While ensemble methods can improve overall accuracy, they often rely on averaging or voting mechanisms that may miss out or even dilute the actual individual strengths of different controllers. As a result, traditional ensemble techniques may thus reduce variance but fail to exploit contextual specialization among controllers. 

In this paper, we reframe the problem of building black-box (AI-based) control ensembles as a contextual monitoring problem. Rather than blending all controller outputs, a runtime monitor identifies and exploits contextual expertise, i.e., based on the current operational context of the ACPS, such as environmental conditions, it determines which controller is safest to deploy. If no controller can be trusted to maintain safety, the monitor, following a Simplex-style strategy \cite{simplex,torfah22}, diverts to a fail-safe that guarantees safety, but potentially at the cost of reduced performance. We provide a formalization of the problem of learning contextual monitors for control ensembles and present
a framework for learning such monitors. Starting with a formal system-level specification defining the safety constraints of an ACPS, our framework enables the learning of contextual runtime monitors that come with formal statistical
guarantees on safety.

\begin{figure}[t]
\includegraphics[scale=0.31]{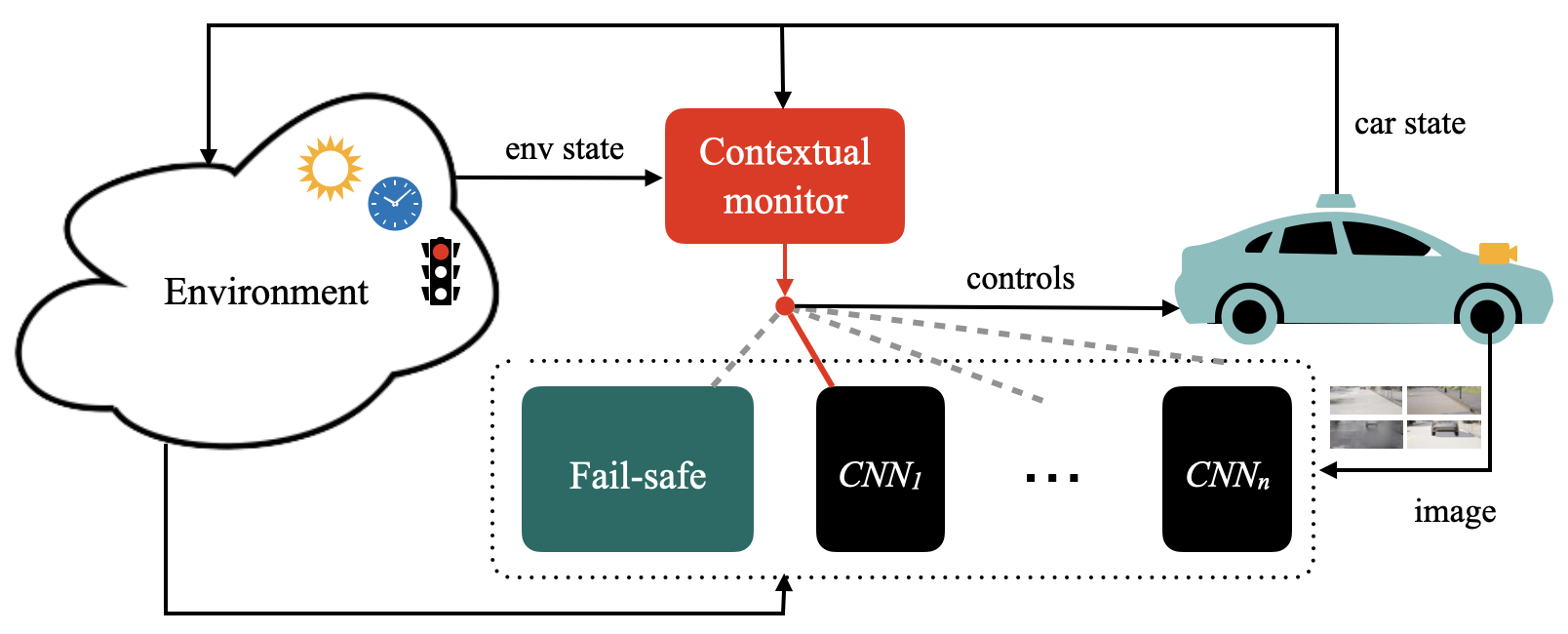}
\caption{An autonomous car equipped with a contextual monitor over an ensemble of CNN-based controllers.}
\label{fig:motexample}
\end{figure}

Consider, for example, the scenario depicted in \Cref{fig:motexample}. An autonomous vehicle is equipped with an ensemble of (black-box) image-based controllers for lane keeping. These controllers could be, for example, realized by an ensemble of convolutional neural networks of different architectures and trained on different data sets.  Depending on how each CNN  was trained and the distribution of its training data, each controller may exhibit specific biases, for instance, due to the presence or absence of particular features during training, or because they were trained on distinct datasets, leading to varying performance across different contexts. 
One controller may perform better in certain weather conditions, another controller may be more robust in certain traffic situations or times of the day. 
The goal is to manage this ensemble via a monitor that learns to identify which controller is most reliable in a given context, and delegate control accordingly. In case the monitor determines that none of the controllers can be trusted to uphold the safety specification in the current context, it triggers a switch to a fail-safe controller, usually realized by a less-optimal but verified control policy, thus ensuring continued safety. A key challenge will be to create a monitor that maintains safety, yet without being too conservative, i.e., resorting to too many unnecessary switches to the fail-safe.



\begin{figure}
    \centering
\includegraphics[width=0.66\linewidth, height= 0.405\linewidth]{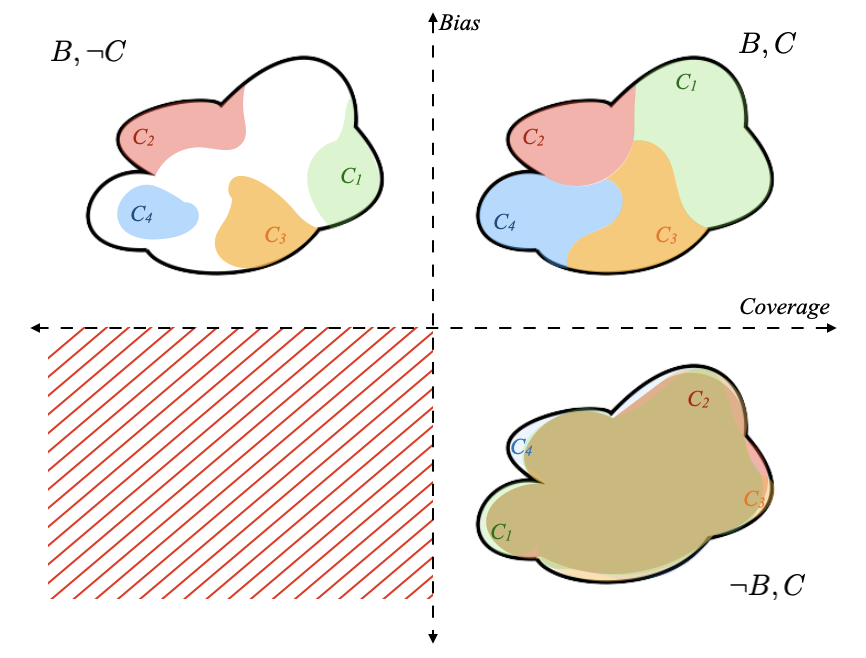}
    \caption{Operational domain bias-coverage spectrum of ensemble-based control.}
    \label{fig:biascover}
\end{figure}

Our contextual approach enables us to leverage the inherent biases in individual controllers, something that stands in contrast to traditional, ensemble methods such as (weighted) averaging over bagged ensembles, boosting, or mixture-of-experts techniques \cite{learningsituationaldriving}. While these methods are effective at reducing overall variance, they typically fail to exploit the specific strengths that individual controllers exhibit in particular operational settings.
This limitation is best understood by considering an ensemble’s behavior relative to the operational domain (OD) of the ACPS. For example, in our earlier scenario, the OD may correspond to the environmental conditions under which the system operates. 
The goal is to construct ensembles that provide robust coverage of the OD,  i.e., ones that can operate safely across most, if not all, contexts within this OD.
As illustrated in \Cref{fig:biascover}, we identify several possible scenarios along this spectrum. In the ideal case, each controller individually offers broad OD coverage, with only minor weaknesses that can be compensated for by others in the ensemble (the $\neg B, C$ case). In such situations, conventional ensemble techniques often perform well at improving accuracy. However, this ideal scenario is rarely guaranteed. In other cases, the ensemble may collectively cover the OD, but individual controllers exhibit biases toward specific regions (the $B, C$ case). In such cases, simple averaging may fail to mitigate biases and can even degrade performance. A worse situation occurs when the controllers are biased and, collectively, fail to provide sufficient OD coverage (the $B, \neg C$ case). In this scenario, our approach can still exploit controller biases to select whichever controller remains safe, if any. 
The final case ($\neg B, \neg C$) represents a poorly trained ensemble in which all methods, contextual or otherwise, will likely lead to safety violations or degraded performance.

 Under this view, the monitor learning problem is thus contextual, one that can be naturally phrased and solved by drawing correspondence with contextual multi-armed bandit problems. Contextual bandits are
 a variation of the multi-armed bandit problem where rewards associated with the arms change at each round depending on the context \cite{NIPS2007_4b04a686,lattimore2020bandit}.  In our setting, the arms represent a set of black-box (ML-based) controllers (e.g., the ensemble of CNN-based controllers from our example above), the context corresponds to the environmental settings in which the system is deployed (e.g., weather, time of day, road features, etc.), sometimes it could also include the system state (e.g., speed), and the rewards are determined by the satisfaction of a system-level specification that defines the system’s safety requirements (e.g., avoiding lane invasions, keeping a certain distance to other vehicles and objects). 
 The goal is to learn a monitor that optimizes performance while maintaining safety, selecting the most suitable controller for a given context, and relying on the fail-safe controller only when necessary.  
From a predefined space of possible monitors, our method learns a monitor whose performance closely matches an optimal monitor within that space, with a guaranteed bound on the error \cite{Das2024ActivePO} quantified in terms of a regret, a standard criterion in bandit literature.

Contextual bandits provide a dynamic and guided approach that allows to continually learn from feedback \cite{BouneffoufRA20}. Our approach contrasts with the naive approach of simply training a classifier to map a context to a controller. The latter is a purely passive learning approach, whereas bandits allow one to adapt to the dynamic nature of  \acps environments and to continually incorporate feedback to improve the system. 
From a practical view, we show that our approach provides significant improvements in performance over methods that use simple, non-contextual approaches. We validate the latter, through a series of experiments, using a case study from the domain of autonomous driving, building on the example above, and demonstrating the efficacy of our approach, highlighting further the importance of contextual learning in building control ensembles for ML-based \acps. 

The rest of the paper is organized as follows. We start by reviewing related work. In \Cref{section:problemformulation}, we introduce and formalize the problem of learning contextual runtime monitors for control ensembles. \Cref{section:theoretical} presents a framework for learning contextual monitors with formal statistical optimality guarantees. In \Cref{sec:experiments}, we provide a thorough experimental evaluation showing the efficacy of our approach in enhancing safety while maintaining a reasonable degree of performance, specifically showing the benefits over non-contextual methods. 


\section{RELATED WORK}

\subsection{Safety Monitoring and Shielding}

The design of runtime monitors has long been, and continues to be, crucial in the development of safe and reliable cyber-physical systems \cite{Bartocci2018,DBLP:conf/rv/TorfahJFS21}.  A key application of runtime monitoring is the validation of design assumptions. This has typically followed a specification-based approach, where design assumptions are captured in formal specifications \cite{DBLP:journals/fmsd/MitschP16, desai-rv17} and directly monitored at runtime. 
In our setting, while we assume access to a general system-level safety specification,  we cannot assume that we can synthesize the monitor directly from this specification, as it is defined over a different space, the context space. Thus, it must be learned using a data-driven approach with respect to this specification \cite{torfah22,10.1145/3716550.3722021}. 
This also stands in contrast to prominent approaches based on shielding~\cite{10.5555/3504035.3504361}. Such approaches typically assume knowledge of system dynamics (e.g., MDP in shielding) along with a conservative environment model. In contrast, our monitors do not make such assumptions, are defined solely over observable features, and do not assume direct monitorability of the specification. A similar argument holds for methods based on barrier certificates \cite{9157966} or safety filters \cite{DBLP:journals/arcras/HsuHF24}. 
Lastly, there is a series of works that address the problem of predictive safety monitoring, e.g., in RL settings to predict the impact of actions on safety \cite{10745554}, and also those in adversarial settings \cite{DBLP:conf/rv/MallickGBD23}. Such approaches can be adapted to our settings to predict contexts to include the aspect of predictiveness, but we keep such a study for future work. Many works also looked into approaches for safe model-predictive control (MPC) (e.g., \cite{10113472,10886383}). While showing safer outcomes, such approaches again require access to a system dynamics model. 
A key distinction to the methods above lies in our focus on addressing the aspect of context-dependent monitoring and how it can help in enhancing safety while balancing performance.

\subsection{Ensemble-based Control}
Ensemble-based control is a widely used technique in control systems, machine learning, and robotics, where multiple controllers (or models) are combined to make decisions or control actions \cite{RAMAKRISHNA2020101760,10886150,Tong2025EnsembleNN,ulgen}. Such ensembles improve robustness by reducing the impact of individual controller errors. Most ensemble-based control methods rely on averaging approaches to smooth out these errors. For example, Ramakrishna et al. demonstrate how adapting weighted control ensembles within a Simplex architecture can improve system performance \cite{RAMAKRISHNA2020101760}. Li et al. show that ensembles of DNN regressors can better account for uncertainty in adaptive cruise control settings~\cite{10886150}. Tong et al. illustrate how ensembles can reduce error bounds in deep model predictive control \cite{Tong2025EnsembleNN}.
Some contextual extensions of weighted averaging have been explored through mixture-of-experts frameworks \cite{learningsituationaldriving}, where a learning-enabled model, such as a neural network, is trained to determine the contribution of each controller based on the observed context. In these approaches, controller outputs are aggregated dynamically according to the weights computed by the model. In contrast, our approach leverages the strengths of each controller without diluting them through aggregation, allowing the system to exploit individual expertise more effectively.

\subsection{Contextual Bandits in Decision-Making}
Contextual bandits are a classical topic in the bandit literature \cite{NIPS2007_4b04a686,lattimore2020bandit}. The logistic bandit setting has received much attention recently in the binary case \cite{FilippiCGS10,FauryAJC22}, see also \cite{CarlssonBJD24} for an application to learning preference orderings. The contextual multiarm bandit model framework has attracted a lot of attention in various applications, from recommender systems and information retrieval to healthcare and finance, because of its stellar practical performance in learning from feedback combined with attractive theoretical properties. See \cite{BouneffoufRA20} for a survey of applications. However, we are not aware of any applications in the domain of interest in this paper.

\section{PROBLEM FORMULATION}

\label{section:problemformulation}

In this section, we formally define the problem of learning contextual monitors. 
The problem is defined over three main ingredients:  \emph{monitor-guided systems}, \emph{safety specifications}, and \emph{contexts}. 
The learning problem is then one where we want to learn a monitor that, when plugged into the monitor-guided systems, will, for any given context, choose a controller that is trusted to maintain the safety specification, up to some statistical boundaries, and decide to switch to a fail-safe if no controller is trusted to be used safely.

We start by formally introducing the notion of {monitor-guided systems},
{safety specification} and {contexts}. 

 \textbf{Notation}
If $V$ is a set of variables that are defined over a domain $\mathbb D$, we define a valuation of $V$ as a function $\nu\colon V \rightarrow \mathbb D$, and write the set of valuations of $V$ as $\mathbb D^V$.
For a set $A\subseteq V$, we define $\nu_A\colon A \rightarrow \mathbb D$ as the valuation resulting from restricting $\nu$ to the variables in $A$. 
Lastly, we use $Z^*$ to refer to the set of finite sequences over elements of a set $Z$, also referred to by the traces over $Z$. 

\subsection{Monitor-Guided Systems}
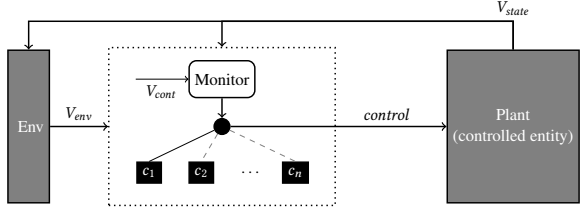
\begin{figure}
    \centering
\scalebox{0.65}{
    \begin{tikzpicture}
    \node[draw, rounded corners, thick, minimum height=0.7cm](M) at (0,0.9) {Monitor};

    \node[left = 1cm of M](cont){};
    
    \path[draw,->](cont) node[below right= 0 and 0.2cm]{$\vcont$} --(M.west);
    
    \node[draw,thick, dotted, minimum height=3cm, minimum width = 4.3cm](mgs)at (0,0){};

    \node[draw,fill=black,thick, below left = 1.2cm and 0.5cm of M, text = white](c1) {$c_1$};

    \node[draw,fill=black,thick,  right = 0.5cm of c1, text = white](c2) {$c_2$};

    \node[right = 0.4cm of c2, text = black](cdots) {$\dots$};

    \node[draw,fill=black,thick, below right = 1.2cm and 0.5cm of M, text = white](cn) {$c_n$};

    \node[draw,circle, fill=black,thick](switch) at (0,0){};

    \path[draw,thick,->] (M) -- (switch);

    \path[draw] (switch) -- (c1.north);

    \path[draw, dashed, gray] (switch) -- (c2.north);

    \path[draw, dashed, gray] (switch) -- (cn.north);

    \node[draw, minimum height= 2.9cm, thick, right = 2.1cm of mgs, fill = gray, text= white, align= center] (P) {Plant\\(controlled entity)};

    \path[draw,->, thick](switch) -- node[above right = 0 and 0.35cm]{$\vctrl$} (P.west);

    \node[draw, minimum height= 2.9cm, thick, left = 1.1cm of mgs, fill= gray, text= white] (E) {~Env~~};

    \path[draw,->, thick](P.north) |-node[above]{}  (0,2) -| (E.north);

    \path[draw,->, thick](E) -- node[above]{$\venv$}(mgs);

    \path[draw,->, thick](P.north) |-node[above]{$\vplant$}(0,2) -| (mgs.north);

    \end{tikzpicture}

}
    \caption{A general view of a monitor-guided system consisting of $n$ controllers managed by a contextual monitor.
    }
    \label{fig:mgs}
\end{figure}

An abstract view of the architecture of systems studied in our paper is depicted in \Cref{fig:mgs}. The system is composed of an environment and a controlled entity. This entity is controlled by a set of controllers $\{c_1, \dots, c_n\}$, which in turn are managed by a monitor. The controls computed by a controller determine how the state of the controlled entity changes.
The role of the monitor is to dynamically adapt the controls of the safest controller at each execution step based on the current observed context. The context can be defined in terms of the state of the controlled entity and that of the environment (or a part thereof). We call the architecture in \Cref{fig:mgs} a \emph{monitor-guided systems (\mgs)}.

Formally, we define an \mgs as a tuple $\mathcal{S}=(\venv, \vsys, \vcont,C, \iota, \pi)$. 
The set $\venv$ is a set of environment variables over which we define a state of the environment. 
The set  $\vsys$ is a set of variables over which we define a state of the controlled entity. 
The set $\vcont \subseteq \vsys \cup \venv$ is a set of context variables. Usually, context variables are associated with current sensor measurements or the state of the plant. W.l.o.g., we assume all variables are defined over the same domain $\mathbb D$.  
The set $C$ is a set of controllers, where $c\in C$ implements a function $c\colon  (\mathbb D^{\vsys\cup {\venv}})^* \rightarrow \mathbb D^{\vsys}$. 
The state $\iota \in \mathbb D^{\vsys}$ is the initial system state. 
Lastly, $\pi\colon \mathbb D^{\vcont} \rightarrow C$ is a monitoring policy that for a context $\xi \in \mathbb D^{\vcont}$ returns a controller  $c\in C$.   

\begin{remark}
We note that we consider  \mgs defined in terms of positional contexts, i.e., the output of a monitor depends only on the current context. Such contexts are common in \acps, e.g.,  operational design domains are usually defined in terms of current environmental conditions, constellation of traffic, etc. \cite{saesae}. 
Contexts defined in terms of a bounded history of events can be reduced to the positional setting. 
\end{remark}

An \mgs follows the concept of the well-known Simplex architecture \cite{simplex}, where a decision module is used at runtime to choose between an optimal, yet not necessarily verified controller (in our case, between several of those), and a verified safe controller.  Assuming one of the controllers in an \mgs is a verified safe-controller, the policy plays the role of a decision module, but one that, based on a certain context defined in terms of environment features and system state, decides which controller out of many optimal controllers should be used to control the system, before deciding to switch to the safe controller.

\textit{Specifications.} We define the semantics of an \mgs $\mathcal S$ by its set of executions traces
$\traces{\mathcal S} =  \{ \tau^1 \dots \tau^k \in (\mathbb D^{\vsys \cup \venv})^* \mid  \tau^1_{\vsys} = \iota \wedge \forall 1<i\leq k.~ \tau^i_{\vsys} = \pi(\tau^{i-1}_{\vcont}) (\tau^{1}_{\vsys} \dots \tau^{i-1}_{\vsys})\} $. 
We define the safety of a system in terms of finite trace specifications \cite{checkingfinitetraces}, i.e.,  a set of finite traces from which the system traces should not deviate. A specification is thus a set $\varphi \subseteq (\mathbb D^{\vsys \cup \venv})^*$.
We say that a trace $\tau \in \traces{\mathcal S}$  satisfies $\varphi$, denoted $\tau \models \varphi$ if and only if $\tau \in \varphi$. We further say, that $\mathcal S$ satisfies $\varphi$, denoted $\mathcal S \models \varphi$, if $\forall \tau \in \traces{S}.~ \tau \models \varphi$.

Putting all concepts together, an example of a safety specification in the scenario of the autonomous car in \Cref{fig:motexample} could be an invariant requiring that there are no lane invasions, and that the car should always keep a certain distance from other objects on the road. A contextual monitor will now have to decide based on an observed context, e.g., weather condition, which controller is safest, in terms of satisfying the specification above. 

Several formalisms exist in the literature to describe finite trace specification (e.g., logics like STL \cite{DBLP:conf/formats/MalerN04}, automata \cite{checkingfinitetraces}). Our approach is agnostic to any specification formalism, as long as specifications are monitorable.

\subsection{Problem Statement} \label{section:problemstatement}
Using the concepts above, our contextual monitor learning problem can now be defined as follows. Let $\mathcal S =(\venv, \vsys, \vcont,C, \iota, ?)$ be an \mgs with a missing monitoring policy, and with $\vsys$ and $\venv$, defined over a domain~$\mathbb D$.  Let further a specification $\varphi \subseteq (\mathbb D^{\vsys \cup\venv})^*$, and $\Pi$ be a set of policies from $\mathbb D^{\vcont} \rightarrow C$. Our goal is to find a policy $\pi \in \Pi$ 
closest to an optimal policy $\pi^*$, i.e., one that optimally selects the safest controller for a given context. 
The distance is measured in terms of the following   \emph{regret} \cite{lattimore2020bandit}:
\begin{align*}
    \max_{\xi \in {\mathbb D^{\vcont}}} |L_{\mathcal S}(\pi(\xi),\varphi) - L_{\mathcal S}(\pi^*(\xi),\varphi)|    \\ \text{ where }   L_{\mathcal S}(\pi(\xi),\varphi) := \prob{\mathcal S[\pi] \not \models \varphi \mid \xi}
\end{align*}  
with $\mathcal{S}[\pi]$ defining the \mgs resulting from replacing ? with $\pi$ in $\mathcal S$, i.e., $\mathcal{S}[\pi]$ follows the behavior determined by $\pi$. The regret is thus the maximal difference between the loss suffered by the optimal controller and the loss of the controller computed by our policy $\pi$   maximized over all contexts.
Our goal is to \emph{minimize} this regret. Note that we consider a quantitative notion of safety defined in terms of the probability of violating a safety specification. Minimizing regret means targeting a monitor that minimizes this probability to closely match that of the optimal one.  

In the next section, we present an algorithm that learns a monitoring policy with bounds on this regret. 

\section{LEARNING CONTEXTUAL MONITORS}
\label{sec:learning_contextual_monitors}

\label{section:theoretical}

Our learning problem is inherently contextual, and can be naturally approached by drawing on techniques from the contextual bandit literature~\cite{lattimore2020bandit}. This enables us to obtain formal regret-minimization bounds as we show in this section.

We restrict ourselves to monitors that model the violation probability using logistic regression \cite{logisticreg}. That is, given a controller $c$ and context $\xi$, we assume that there is some unknown vector $\theta_c$ such that the probability that $c$ violates the specification, given $\xi$, can be written as $\prob{Y=1|c,\xi} = \sigma(\theta_c^\top \xi)$
where $\sigma(.)$ is the logistic function, and $Y$ is a Bernoulli variable representing whether a violation has occurred. We also restrict ourselves to bounded monitors and bounded contexts, technical assumptions which are standard in the bandit literature (see \cite{Das2024ActivePO} and references therein).

{ Following the logistic regression model, we can then rewrite the regret as defined in our problem statement by substituting the loss with $ \sigma(\theta_c^\top \xi)$. Solving the problem is then to find a policy that minimizes the following regret:}
\begin{align*}
\max \limits_{\xi \in \mathbb D^{\vcont}} |\sigma(\theta_{\pi(\xi)}^\top \xi) - \sigma(\theta_{\pi^*(\xi)}^\top \xi) |. 
\end{align*}
Learning an optimal policy $\pi$  resorts now to solving a minimization problem for the vectors $\{\theta_{c}\}_{c\in C}$. In the following, we define a matrix $\theta$ whose rows are the vectors $\theta_c$, and we describe a learning approach based on contextual bandits to learn this matrix. This overall process is shown in \Cref{alg:active_learning}. 
In our setting, a learner interacts with the system over a sequence of $T$ rounds. At each round~$t$, the learner
selects a context $\xi$ and a controller $c_t$, and runs this controller on the system in that context. 
Before moving on to the next round, the learner observes the outcome $Y_t$. In our setting, $Y_t$ is a binary variable that tells whether the system using the chosen controller violated the safety specifications $\varphi$. 
At the end of each iteration $t$, the row $\theta_c$ of the matrix $\theta$ is updated with the values $\theta_{t,c}$, which are computed for a selected controller $c$ as the maximum-likelihood estimate (MLE):
\begin{align*} 
   \theta_{t, c} \in \argmax_{\theta'_c \in {\mathbb{R}^{|\vcont|}}} \sum_{j=1}^t \mathbb{1}_{\left(c_j = c' \right)}  &\Big( Y_j \log \sigma(\theta_c'^\top \xi_j) + \nonumber  \\ 
   & (1-Y_j)\log(1- \sigma(\theta_c'^\top \xi_j)) \Big) \nonumber
\end{align*}
After  $T$ rounds, the learner outputs a monitor $\pi_T$ corresponding to the matrix $\theta$, whose rows are the vectors $\theta_{T,c_1},\dots, \theta_{T,c_{|C|}}$. 
In the following, we provide details about the individual steps.

\paragraph{\textbf{Select context and controller (Line 3)}}

The key component of our approach is how to decide which controller to test over what context during the learning phase. We do this using the value of an uncertainty metric based on the Hessian of the negative log-likelihood: 
\begin{align*}
    \Hess_{t}(\theta_{c, t}) = \sum_{s=1}^t \mathbf{I}_{(c_s = c)} \ \dot{\sigma}(\theta_{c, t}^\top \xi_s) \xi_s \xi_s^\top 
\end{align*} 
where $\dot{\sigma}(.)$ is the derivative of $\sigma(.)$.
Intuitively, the Hessian indicates the curvature while moving in a particular direction. A higher curvature indicates more sensitivity to changes in that direction, which is linked to more uncertainty because small changes lead to unpredictable results. This notion of uncertainty is inspired by work done on logistic bandits~\cite{FilippiCGS10,faury2020improved,kveton2020randomized}. The technical details can be found in the Appendix.

 Choosing a controller in a round can now be done in many ways. A prominent way 
is to randomly sample a context first, according to a given distribution, and then choose the controller with the highest uncertainty. 
Another method that provides better theoretical upper bounds on regret, and which we show to be effective in our experiments, is to choose the controller with the highest uncertainty over all contexts. In \Cref{theorem:theorem1}, we show the order of these bounds. This is an adaptation of the sampling rule of \cite{Das2024ActivePO} to our setting.

At a high level, the computation in round $t$ can be done using the current MLE values $\theta_{t,c}$ for each controller $c$ and the corresponding Hessian of the negative log-likelihood to measure the epistemic uncertainty about a controller's safety in a given context. Since initially we do not have information about the values $\theta_{t,c}$, the implementation runs the system once with each controller in a random context before using the uncertainty values. The complete mathematical formulation can be found in the Appendix.

\paragraph{\textbf{Evaluate a controller (Line 4)}} A selected controller $c_t$ is evaluated over the context $\xi_t$, with respect to the specification $\varphi$. Evaluation is done by running a simulation where the system is restricted to only using the controller $c_t$, and computing $Y_t$, which captures whether the specification $\varphi$ was violated or not. 

\paragraph{\textbf{Update monitor and uncertainty of controllers (Line 6)}}  
The monitor is retrained over a new set $\mathcal{D}_t$ that expands on $\mathcal{D}_{t-1}$ with new data collected from evaluating the controller (Line 4). The uncertainty update can be performed using the Sherman-Morrison formula \cite{shermanmorrison}. A full mathematical walk-through is kept to the Appendix.

\begin{algorithm}[t]
\caption{Contextual monitor learning}\label{alg:active_learning}
\begin{algorithmic}[1]
\REQUIRE Bandit algorithm \textbf{Alg}, specification $\varphi$.
\STATE Initial empty dataset $\mathcal{D}_0 = \{\}$
\FOR{$t=1,..., T$}
\STATE  $\xi_t, c_t = \textbf{Alg}.\text{select}(\mathcal{D}_{t})$.
\STATE $Y_t=\text{run}(c_t, \xi_t,\varphi)$.
\STATE  $\mathcal{D}_{t} := \mathcal{D}_{t-1} \bigcup \left\{(\xi_t, c_t, Y_t)\right\}$.
\STATE $\pi_t$ = $\textbf{Alg}.\text{update}(\mathcal{D}_{t})$.
\ENDFOR
\STATE return $\pi_T$ 
\end{algorithmic}
\end{algorithm}

By applying \Cref{alg:active_learning} for a sufficient number of iterations, we obtain a monitor that closely approximates the optimal monitor. Formally, we have the following result, which upper bounds the regret by a quantity that converges to $0$ with an increasing number of rounds. The proof is given in the Appendix.

\begin{theorem}
    Using \Cref{alg:active_learning}, selecting the controller and context with the highest uncertainty, we have that the regret is bounded by $\mathcal{O}(\sqrt{\log(T)^2/T})$. 
    \label{theorem:theorem1}
\end{theorem}

\begin{remark}
The reason to sample contexts considering the uncertainty in Alg.~\ref{alg:active_learning} is to deal with the problem of having a lower bound on the suboptimality gap. This arises when the number of rounds $T$ is much lower than the number of contexts $|\mathbb D^{\vcont}|$ (see \cite{Das2024ActivePO} for details). 
\end{remark}

\section{EXPERIMENTAL EVALUATION}

\label{sec:experiments}

We assess the efficacy of our proposed approach by investigating the following research questions: 
\begin{itemize}
    \item {\bf RQ1: Sanity check.} Does a monitor learned using our approach indeed select the controller best for a context?
    \item {\bf RQ2: Comparing to other baselines.} How does our approach compare to other ensemble methods, like common weighted average and mixture-of-experts methods? 
    \item {\bf RQ3: Active bandits vs passive learning.} Does our active learning approach result in more accurate monitors than simple one-shot passive learning approaches?
    \item {\bf RQ4: Simplex vs multi-Simplex.} What is the impact of larger control ensembles on the accuracy of monitors? 
\end{itemize}

We address these research questions by evaluating our approach on two scenarios from the domain of autonomous driving. We first describe the experimental setup, including details about implementation and scenarios, and then present our results.

\subsection{Framework and Implementation}
\label{sec:expsetup}
The workflow of our implementation is shown below.

\begin{figure}[h]
\centering

\scalebox{0.8}{
    \begin{tikzpicture}
\centering
    \node[draw, rounded corners, thick, minimum height=1cm](learner) at (0,0.5) {\begin{tabular}{c}Learner \end{tabular}};
    
    \node[draw,thick, minimum height=1.5cm, left = 1.2cm  of learner](sampler) {Sampler};

    \node[draw,thick, dotted, minimum height=2cm, minimum width = 2.9cm, left = 0.9cm of learner](scenic) {};
    \node[above = 0.1cm of scenic](scenic_text) {\textsc{Scenic}\cite{scenic}}; 

    \node[draw,thick, minimum height=1.5cm, right = 1.2cm  of learner](evaluator) {Evaluator};
    
    \path[thick,<-]  ($(sampler.west)+(0,+0.3)$) edge node[above] {env} ($(sampler.west)+(-1.,+0.3)$) ;
    \path[thick,<-]  ($(sampler.west)+(0,-0.5)$) edge node[above] {sys} ($(sampler.west)+(-1.,-0.5)$) ;
    \path[thick,<-]  ($(learner.north)$) edge node[left] {param} ($(learner.north)+(0,+0.5)$) ;
    \path[thick,<-]  ($(evaluator.north)$) edge node[left] {$\varphi, n$} ($(evaluator.north)+(0,+0.5)$) ;
    \path[thick,->]  ($(sampler.east)$) edge node[above] {contexts} ($(learner.west)$) ;
    \path[thick,->]  ($(learner.east)+(0,+0.1)$) edge node[above] {$c, \xi$} ($(evaluator.west)+(0,+0.1)$) ;
    \path[thick,<-]  ($(learner.east)+(0,-0.2)$) edge node[below] {$r$} ($(evaluator.west)+(0,-0.2)$) ;
    \path[draw,thick,<->] ($(evaluator.south)$) |- node[above] {}($(evaluator.south)+(0,-0.5)$) |- node[above] {} ($(scenic.south)+(0,-0.5)$) |- node[above] {}  ($(scenic.south)$) ;

    \end{tikzpicture}
}
\end{figure}

The \emph{Learner} implements the main loop in \Cref{alg:active_learning}. 
The parameters of the learner include the set of controllers ${C}$, the total number of rounds $T$, and the number of rounds $e$ the Learner collects data in before retraining the monitor.
While \Cref{alg:active_learning} 
chooses in each iteration the context $\xi$ and the controller $c$ with the highest uncertainty, in practice, to avoid computation overhead, we sample a large number of contexts, then, the learner chooses the context and controller with the highest uncertainty. 
Sampling the contexts and computing the reward is done by the Sampler and Evaluator, respectively. 

The \emph{Sampler} has access to a system model.  In our implementation, we realize the modeling and sampling using the scenario description language \textsc{Scenic}\xspace \cite{scenic}. A scenic program in our setting will define a distribution over the context space. 
Using  \textsc{Scenic}\xspace, we generate random scenarios in which we simulate our system. Each sampled scenario defines a unique context. 
Once a context $\xi$ and $c$ are selected, they are forwarded by the Learner for evaluation. 

The \emph{Evaluator} then runs the system in $\xi$  using $c$ and evaluates the execution to obtain a (binary) reward $r$. The reward is determined by validating the satisfaction of a safety specification $\varphi$ in a simulation of $n$ steps. The learner uses the reward to update the training data and, in turn, learns a new model. In our setup, the evaluator is also realized using the \textsc{Scenic} execution engine.

\begin{figure*}[t!]
    \centering
    \includegraphics[width=0.20\textwidth]{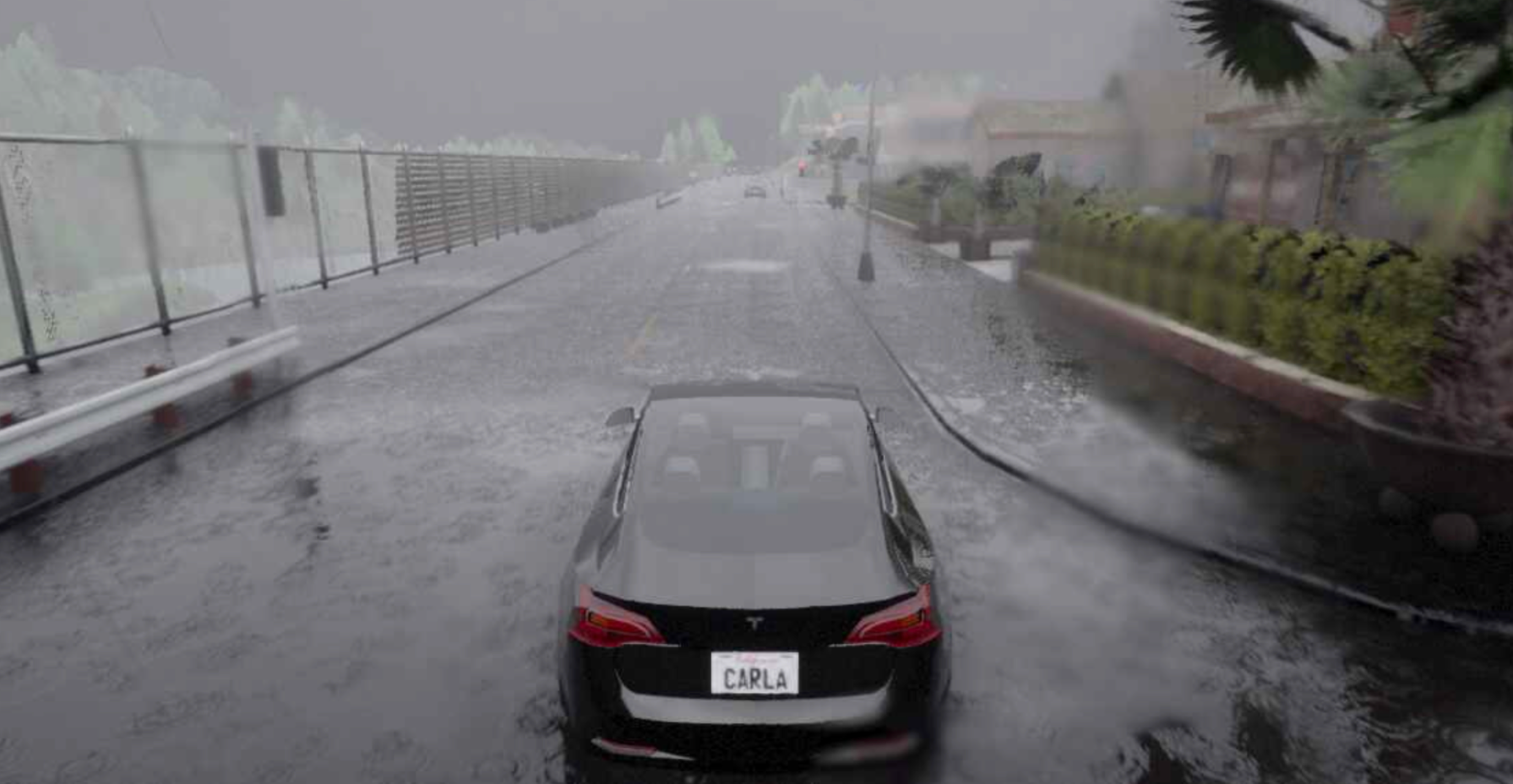}
    \includegraphics[width=0.20\textwidth]{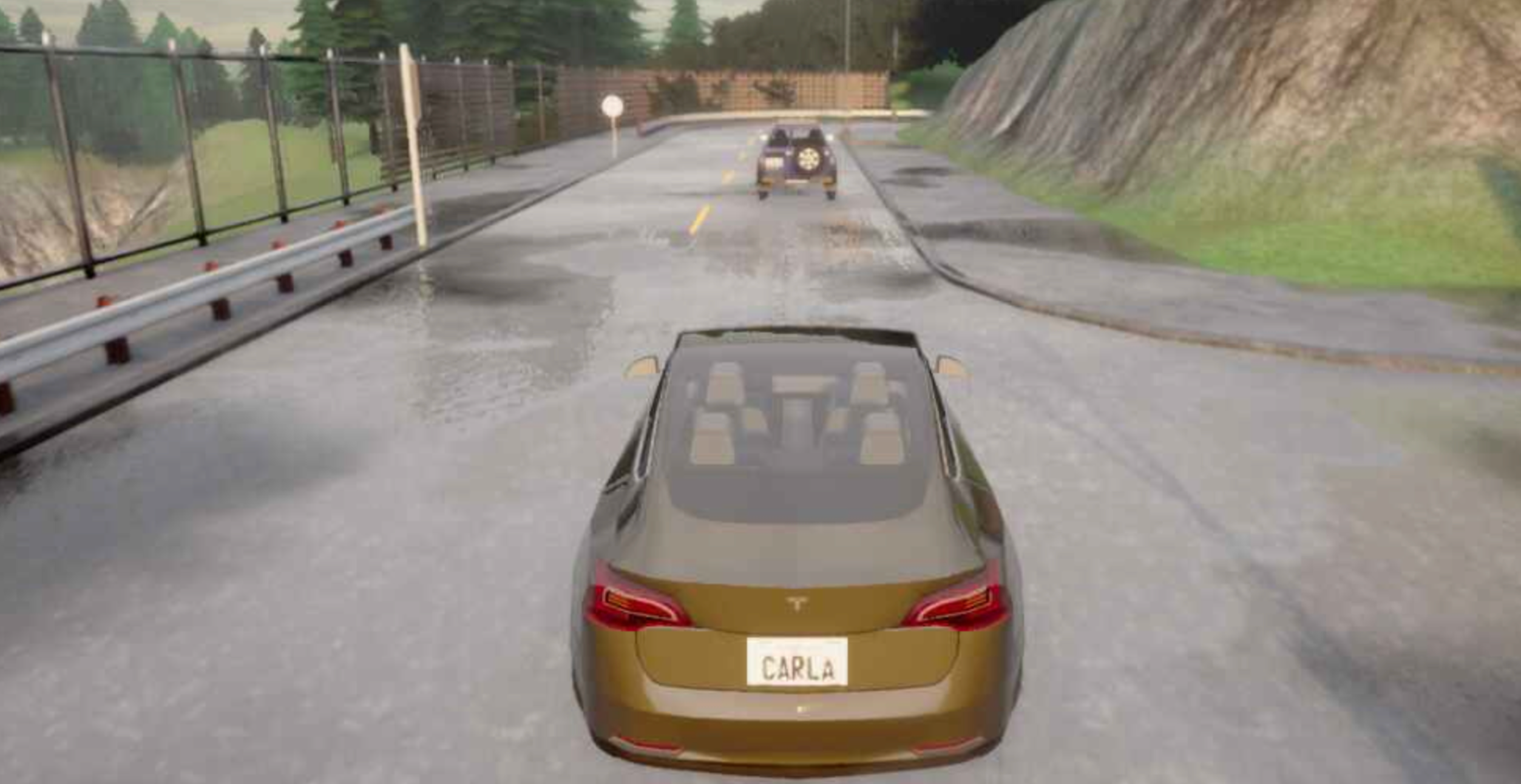}
    \includegraphics[width=0.20\textwidth]{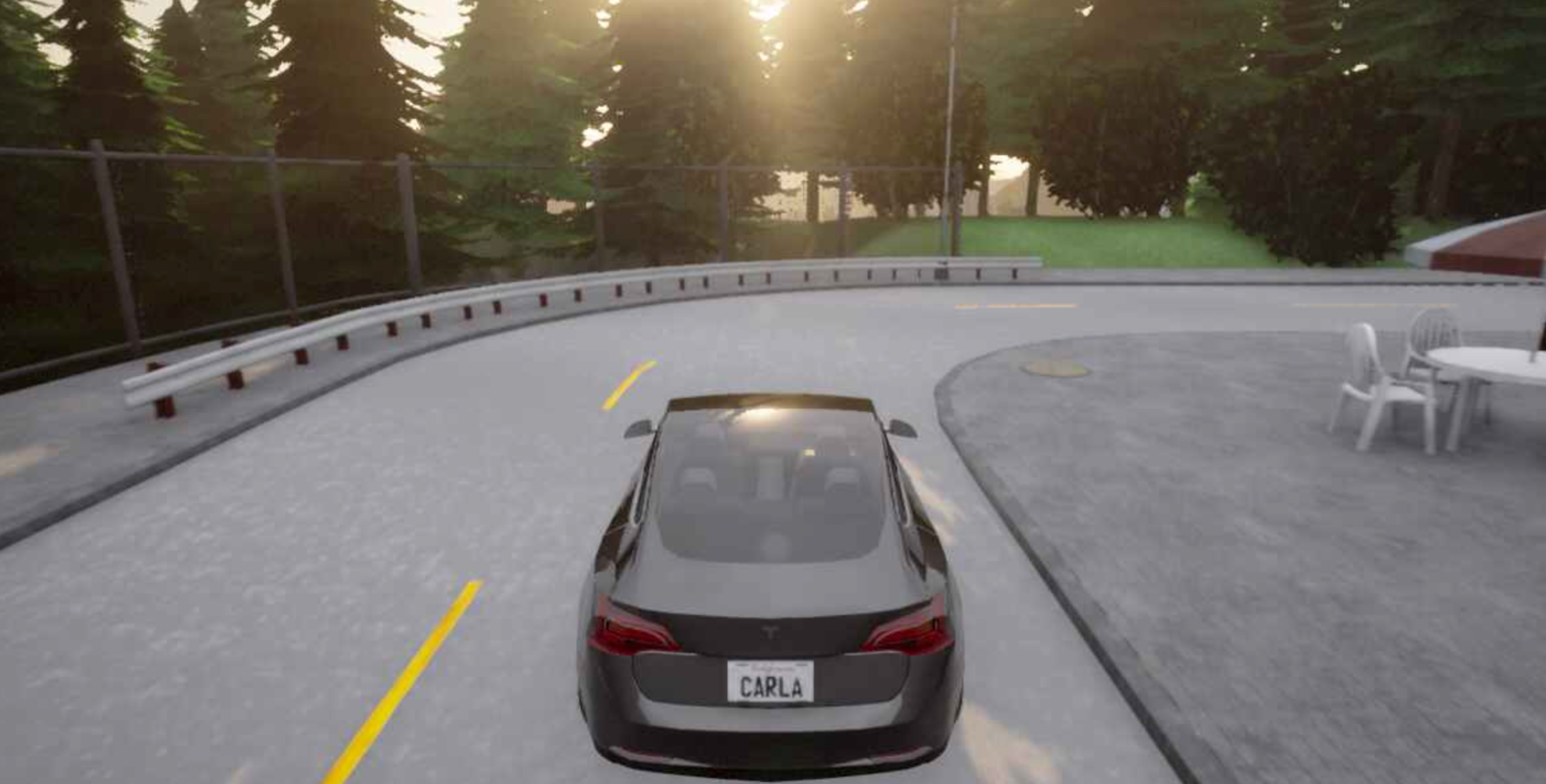}
    \caption{Three instances of Scenario 1, with different weather conditions, distance between cars, and road types.  }
    \label{fig:scenario1}
\end{figure*}

\begin{figure*}
    \centering
    \includegraphics[width=0.20\textwidth]{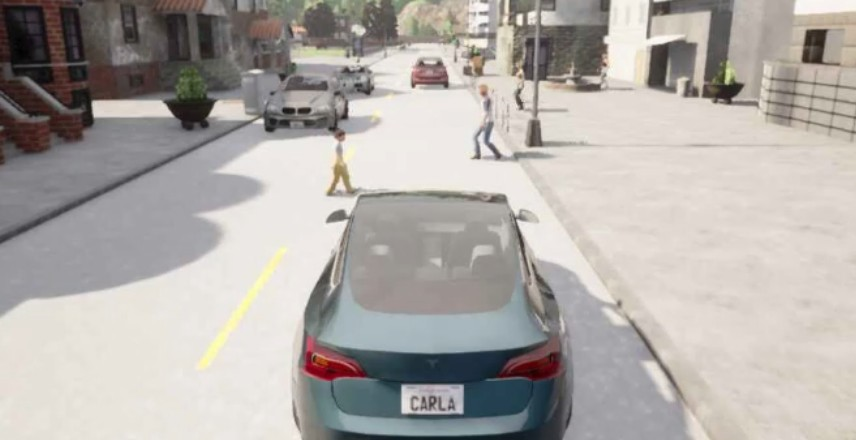}
    \includegraphics[width=0.20\textwidth]{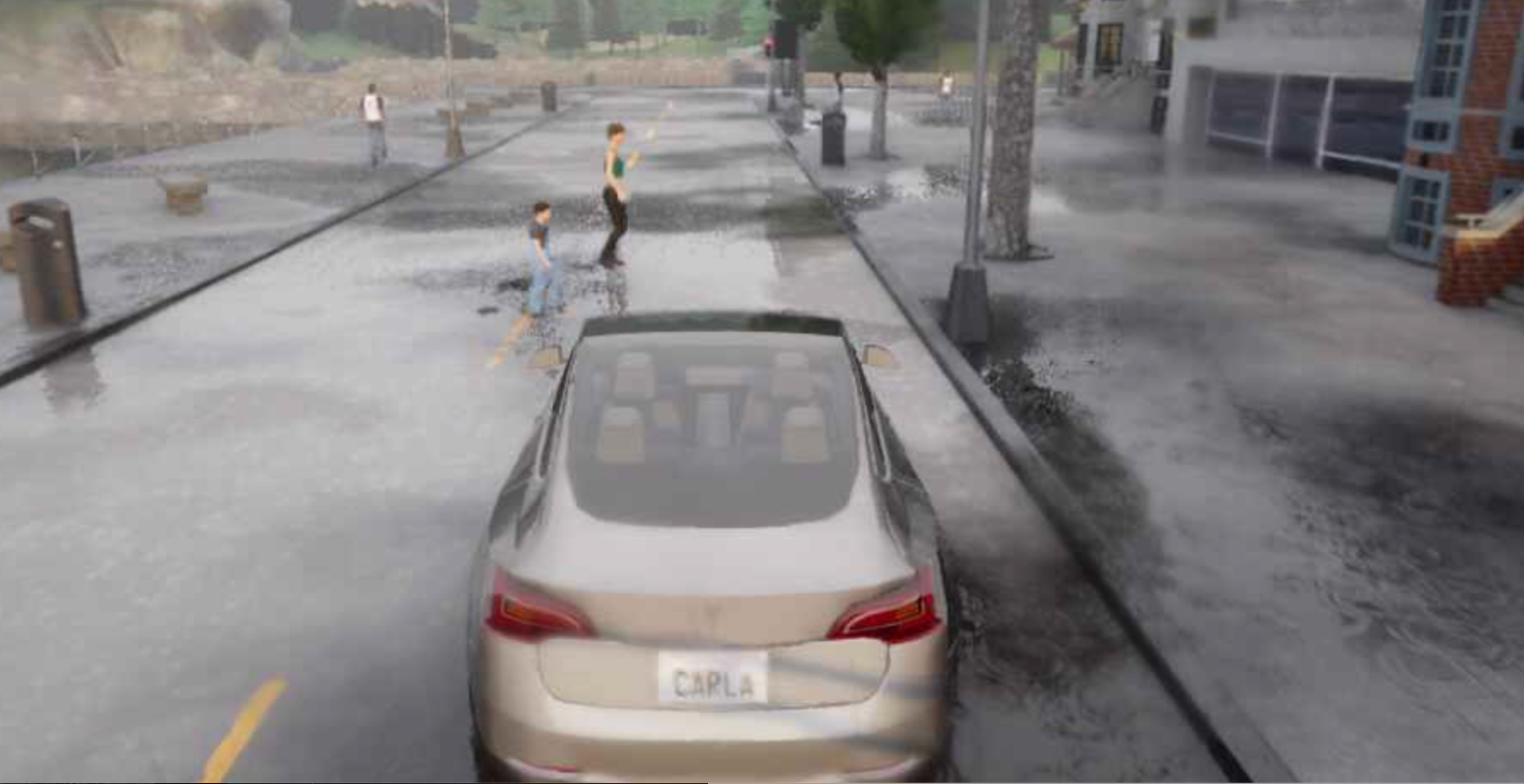}
    \includegraphics[width=0.20\textwidth]{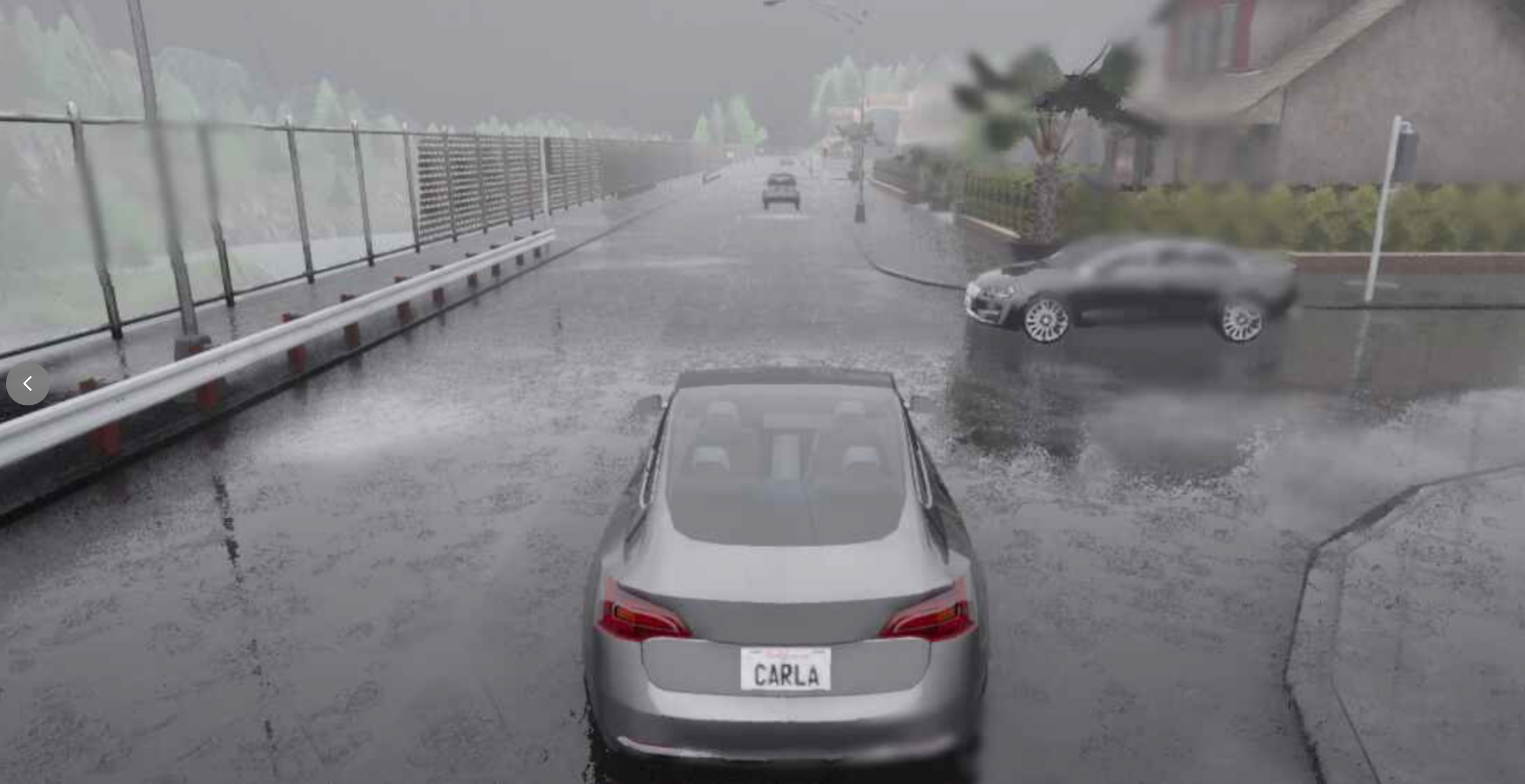}
    \caption{Three instances of Scenario 2, with different weather conditions, and different pedestrians and cars configurations.}
    \label{fig:scenario2}
\end{figure*}

\subsection{Scenarios} \label{section:scenarios}
 Our scenarios were inspired by real-world problems of how to safely deploy vision-based models for autonomous steering and collision avoidance \cite{10.1007/978-3-030-53288-8_6,DBLP:conf/aaai/Chen0Z0W24,7995975}. Our scenarios are simulated using the open source high-fidelity \textsc{Carla}\xspace simulator \cite{carla}. We use \textsc{Scenic}'s interface to \textsc{Carla} to generate random scenarios by sampling from the context space.  
In the following, we provide details for our two scenarios, particularly the controllers, context space, and specification used.  \Cref{fig:scenario1,fig:scenario2} show sample instances for both scenarios.

\subsubsection{Scenario 1 (Autonomous steering)}
An autonomous car is driving in a city. The car is equipped with a set of controllers that compute the steering angle of the car based on images received from a camera. Each controller is composed of a CNN that computes the cross-track error (CTE) to the lane centerline, and a PID controller that uses the CTE to compute the steering angle. 
Additionally, the car includes a fail-safe controller based on the internal follow-lane behavior of Scenic's interface to \textsc{Carla}, providing non-optimal but safe fallback control.

\textit{CNN-based controllers.} 
We trained a variety  of controllers.  We will give details in each experiment on the controllers and the specific datasets used. To generate training data for the controllers, we ran simulations for each of the 14 weather presets with another car positioned at a random distance ahead of the ego car. We collected images from simulations using a camera installed in the front of the car. Each image was associated with the CTE value. We collected $\sim$90K data instances to train each controller. 

\textit{Specification.} 
The safety property enforces lane keeping: the vehicle must not leave its lane for more than 30 consecutive simulation steps.

\textit{Contexts.} We assume the presence of another car. The context space is defined over the following features, which may change: the weather, the time of day, road features such as being on a straight road or at an intersection, and the distance to the other cars ahead (see examples in \Cref{fig:scenario1}). Each context in our scenario is thus represented by a tuple $(w,t,i,d)$, where $w$ represents the weather, $t$ the time of day, $i$ the road type, and $d$ the distance to the vehicle ahead if present. For weather and time of day, we use one of the 14 weather-time configurations predefined in \textsc{Carla}.  The road type $i$ is a binary value, indicating whether the car is at an intersection ($i=1$) or not ($i=0$). The distance $d$ is a real number s.t. $d\in (0,50] \cup \{100\}$. If the other vehicle is not present, or if $d>50$, we assume that the vehicle ahead is too far away to consider it part of the context ($d=100$).  Note that, 
while $w$ and $t$ are fixed for each simulation, $i$ and $d$ can change depending on the movement of both vehicles.
In our experiments, we discretize the distance $d$ into 5 clusters, so we explore a set of 140 different contexts.

\subsubsection{Scenario 2 (Dynamic urban environment)}

This scenario evaluates the collision avoidance ability in a dynamic urban environment, as illustrated in \Cref{fig:scenario2}. In this case, the scenario involves interactions with other agents on the road, including pedestrians and other vehicles. The CNN-controllers not only compute the CTE but also the target speed, which are passed to the PID controller to determine the corresponding throttle or braking intensity. The car is also equipped with a fail-safe controller, with additional brakes when the distance to another car or pedestrian ahead is lower than a given threshold.

\textit{CNN-based controller.} 
Similar to Scenario 1, we trained a variety of controllers over different datasets. To collect data, we ran simulations for each of the 14 weather presets with other cars or pedestrians positioned at a random distance ahead of the ego car. The training images were also taken from simulations. Each image was associated with the CTE value and the current speed. We collected $\sim$90K data instances to train each controller. 

\textit{Specification.} We focus on the ability to perform lane keeping and avoid collisions. The safety specification used is thus one that requires at no time for the car to crash or to leave its lane for more than 30 simulation steps. 

\textit{Contexts.} Each context is represented by a tuple $(w,t,i,d_c, d_p)$, where $w$ represents the weather, $t$ the time of day, and $i$ the road type as same as defined in Scenario 1. The variable $d_c$ represents the distance to the nearest visible vehicle if present, while $d_p$ represents the distance to the nearest visible pedestrian if present. 
Again, $w$ and $t$ are fixed for each simulation, while $i$, $d_c$, $d_p$ can change depending on the movement of the agents in the simulation. After discretizing $d_c$ and $d_p$ in 5 clusters each, we have a total of 700 contexts to explore.\\

\noindent \emph{The algorithm is implemented in Python, and all the experiments were conducted on a machine with a 2.9 GHz 32-Core CPU, 576GB of RAM, and an NVIDIA Tesla T4 GPU.}

\subsection{RQ1: Sanity check} \label{section:exp1}
In this experiment, we verify that a monitor learned using our approach indeed selects the best controller for a context. 
We trained 4 biased controllers on Scenario~1.
One controller is trained on contexts $\xi_1 = (w,t,i,d)$ where the car ahead is close, ($d \in [5,10]$) and the weather/time $(w,t)$ is given by the  \textsc{Carla} preset \emph{ClearNoon}. 
The other three controllers are trained on contexts without another car ahead, $\xi_2 = (w,t,i,100)$, and with weather/time contexts $(w,t)$ from $W = \{$\emph{ClearSunset, HardRainNoon, HardRainSunset}$\}$, respectively. 
In our learning process, we train the monitor only in contexts of the form $\xi_1$ or $\xi_2$. 
After learning, we run a series of random simulations, but also exclusively on instances coherent to $\xi_1$ or $\xi_2$. 
In each simulation, we check if the monitor chooses the expected controller $c_\xi$ when observing a context $\xi$. We used our approach to train three monitors. The hyperparameters chosen were $T=800$ (total
number of rounds), $e=25$ (number of rounds the Learner collects
data before retraining the monitor), and $n=300$ (number of simulation steps, see also \Cref{sec:expsetup}). The monitors were evaluated over 100 simulations every 50 rounds.

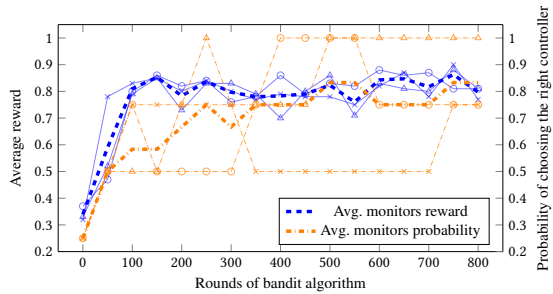
\begin{figure}
\centering
  \scalebox{0.65}[0.65]{
        \begin{tikzpicture}
    \begin{axis}[
            xlabel={Rounds of bandit algorithm},
            ylabel={Average reward},
            xmin=-50, xmax=850,
            ymin=0.2, ymax=1.05,
            axis y line*=left,
            xtick={0,100,200,300,400,500,600,700,800},
            ytick={0.2,0.3,0.4,0.5,0.6,0.7,0.8,0.9,1.0},
            legend pos=south east,
            ymajorgrids=false,
            height=.35\textwidth,
            width= 0.6\textwidth
        ]
        \addplot[color=blue, mark=x,opacity=0.5] coordinates {(0,0.32)(50,0.78)(100,0.83)(150,0.85)(200,0.80)(250,0.84)(300,0.80)(350,0.77)(400,0.79)(450,0.78)(500,0.78)(550,0.75)(600,0.82)(650,0.87)(700,0.78)(750,0.90)(800,0.77)};
        \addplot[color=blue, mark=o,opacity=0.5] coordinates {(0,0.37)(50,0.47)(100,0.80)(150,0.86)(200,0.82)(250,0.84)(300,0.76)(350,0.78)(400,0.86)(450,0.79)(500,0.83)(550,0.82)(600,0.88)(650,0.86)(700,0.87)(750,0.81)(800,0.81)};
        \addplot[color=blue, mark=triangle,opacity=0.5] coordinates {(0,0.33)(50,0.52)(100,0.79)(150,0.85)(200,0.73)(250,0.83)(300,0.83)(350,0.79)(400,0.70)(450,0.80)(500,0.86)(550,0.71)(600,0.83)(650,0.81)(700,0.80)(750,0.88)(800,0.81)};
        \addplot[color=blue,style=dashed,line width=2pt] coordinates {(0,0.34)(50,0.59)(100,0.81)(150,0.853)(200,0.783)(250,0.837)(300,0.797)(350,0.780)(400,0.783)(450,0.790)(500,0.823)(550,0.760)(600,0.843)(650,0.847)(700,0.817)(750,0.863)(800,0.797)};

        \legend{,,,Avg. monitors reward}
        \addlegendimage{color=orange,style=dashdotted,line width=2pt}
        \addlegendentry{Avg. monitors probability}
    \end{axis}
    \begin{axis}[
        xmin=-50, xmax=850,
        ymin=0.2, ymax=1.05,
        axis y line*=right,
        axis x line=none,
        ytick={0.2,0.3,0.4,0.5,0.6,0.7,0.8,0.9,1.0},
        ylabel=Probability of choosing the right controller,
        legend pos=south east,
         height=.35\textwidth,
         width= 0.6\textwidth
    ]
    
        \addplot[color=orange,style=densely  dashdotted, mark=x,opacity=0.7,mark options={solid}] coordinates {(0,0.25)(50,0.5)(100,0.75)(150,0.75)(200,0.75)(250,0.75)(300,0.75)(350,0.5)(400,0.5)(450,0.5)(500,0.5)(550,0.5)(600,0.5)(650,0.5)(700,0.5)(750,0.75)(800,0.75)};
        \addplot[color=orange,style=densely  dashdotted, mark=o,opacity=0.7,mark options={solid}] coordinates {(0,0.25)(50,0.5)(100,0.75)(150,0.5)(200,0.5)(250,0.5)(300,0.5)(350,0.75)(400,1)(450,1)(500,1)(550,1)(600,0.75)(650,0.75)(700,0.75)(750,0.75)(800,0.75)};
        \addplot[color=orange,style=densely  dashdotted, mark=triangle,opacity=0.7,mark options={solid}] coordinates {(0,0.25)(50,0.5)(100,0.5)(150,0.5)(200,0.75)(250,1)(300,0.75)(350,0.75)(400,0.75)(450,0.75)(500,1)(550,1)(600,1)(650,1)(700,1)(750,1)(800,1)};
        \addplot[color=orange,style=dashdotted,line width=2pt] coordinates {(0,0.25)(50,0.5)(100,0.5833333333333334)(150,0.5833333333333334)(200,0.6666666666666666)(250,0.75)(300,0.667)(350,0.75)(400,0.75)(450,0.75)(500,0.833)(550,0.833)(600,0.75)(650,0.75)(700,0.75)(750,0.833)(800,0.833)};

    \end{axis}
\end{tikzpicture}
}
\caption{Results for RQ1: Sanity Check. 
} \label{fig:sanitycheck}
\end{figure}

The results are presented in  \Cref{fig:sanitycheck}. In the initial learning rounds, the monitor chooses a controller randomly, so we expect the initial average reward (average satisfaction of the specification) to be low. With each round, the average reward grows. It stabilizes after around 200 iterations with rewards around 0.8 (solid, blue lines). 
We also compare the performance of the monitors against the true probability of the bandit algorithm selecting the correct controller for each context (orange, dashed lines). Note that the observed probabilities may sometimes be lower compared to the higher quality of the monitors.
For example, for one monitor, the probability was 0.75. A deeper analysis revealed that the algorithm correctly identified the expected controllers for highly distinguishable contexts of \textit{ClearNoon} (with a car close) and \textit{ClearSunset}. However, for the contexts of \textit{HardRainNoon} and \textit{HardRainSunset}, which are highly similar, the corresponding controllers performed with nearly equal quality. 
As a result, the bandit algorithm may not consistently differentiate between them, and the probability of choosing the right controller varies in our experiment between 0.5 and 1.0. 
Consequently, a limitation arises: identifying contexts and learning from them highly depends on the expressiveness of the available observable features.
We conclude that our approach successfully learns to select a suitable controller for the right context if such is available. 
If not, the fail-safe will be triggered.
{ Note, that  we assume at no time access to an optimal monitor. Our results show consistent improvement across learning iterations, reflecting the theoretical guarantees as presented in \Cref{theorem:theorem1}.}

\subsection{RQ2: Comparing to other baselines} \label{section:exp2}

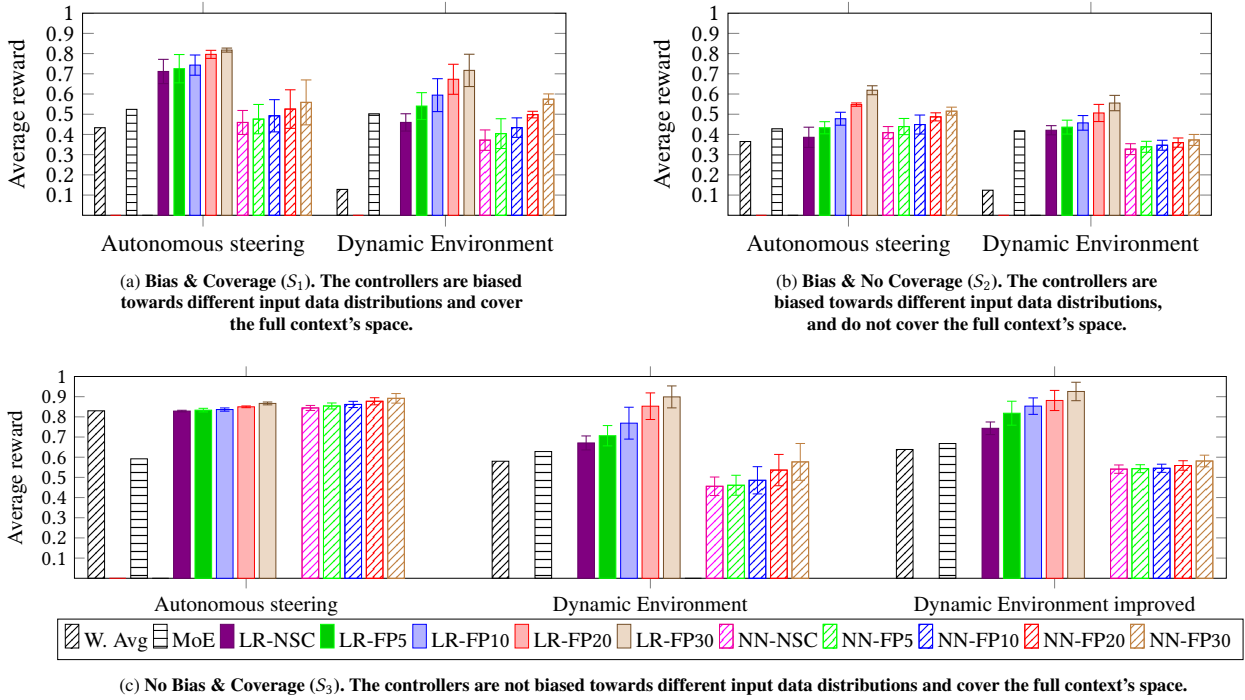
\begin{figure*}
\centering
\begin{subfigure}[b]{0.49\textwidth}
    \scalebox{0.65}{
    \begin{tikzpicture}
        \centering
    \begin{axis}[
            symbolic x coords={{Autonomous steering}, {Dynamic Environment},  
            },
            enlarge x limits={abs=0.3\linewidth},
            width=1.5\linewidth,
            height=0.6\linewidth,
            xtick=data,
            ylabel={Average reward},
            ymin=0.0, ymax=1,
            ytick={0.1,0.2,0.3,0.4,0.5,0.6,0.7,0.8,0.9,1.0},
            ybar,
            bar width=6pt,
            legend image code/.code={
                \draw [#1] (0cm,-0.1cm) rectangle (0.2cm,0.25cm); },
            legend style={at={(0.5,-0.2)},anchor=north,legend columns=-1},
        ]
        \addplot[pattern=north east lines] coordinates {({Autonomous steering}, 0.43333333)(Dynamic Environment, 0.128)
        }; 
        \addplot coordinates {({Autonomous steering},0)(Dynamic Environment, 0)
        }; 
        \addplot[pattern=horizontal lines] coordinates {({Autonomous steering}, 0.524)(Dynamic Environment, 0.502)
        }; 
        \addplot coordinates {({Autonomous steering},0)
        }; 
        \addplot+[error bars/.cd, y dir=both,y explicit] coordinates {({Autonomous steering}, 0.7115253375542604) +- (0.07,0.06)({Dynamic Environment},0.4595) +- (0.0623,0.0428)
        }; 
        \addplot+[error bars/.cd, y dir=both,y explicit] coordinates {({Autonomous steering}, 0.7253152984627191) +- (0.08,0.07)({Dynamic Environment},0.5404) +- (0.0005,0.0663)
        }; 
        \addplot+[error bars/.cd, y dir=both,y explicit] coordinates {({Autonomous steering}, 0.743021438)+-(0.01,0.05)({Dynamic Environment},0.5944) +- (0.0236,0.0814)
        }; 
        \addplot+[error bars/.cd, y dir=both,y explicit] coordinates {({Autonomous steering}, 0.7961398661774228)+-(0.04,0.02)({Dynamic Environment},0.6729) +- (0.0768,0.0743)
        }; 
        \addplot+[error bars/.cd, y dir=both,y explicit] coordinates {({Autonomous steering}, 0.81729544083)+-(0.05,0.01)({Dynamic Environment},0.7169) +- (0.0813,0.0799)
        }; 
            \addplot+[
        magenta,
        fill=white,
        postaction={
            pattern=north east lines, pattern color=magenta
        },error bars/.cd, y dir=both,y explicit] coordinates {({Autonomous steering},0.4593) +- (0.1177,0.0593)({Dynamic Environment},0.3718) +- (0.0482,0.0505)
            }; 
            \addplot[
        green,
        fill=white,
        postaction={
            pattern=north east lines, pattern color=green
        },error bars/.cd, y dir=both,y explicit] coordinates {({Autonomous steering},0.4758) +- (0.1177,0.0725)({Dynamic Environment},0.4040) +- (0.0236,0.0737)
            }; 
            \addplot[
        blue,
        fill=white,
        postaction={
            pattern=north east lines, pattern color=blue
        },error bars/.cd, y dir=both,y explicit] coordinates {({Autonomous steering},0.4924) +- (0.1177,0.0797)({Dynamic Environment},0.4337) +- (0.0452,0.0483)
            }; 
            \addplot[
        red,
        fill=white,
        postaction={
            pattern=north east lines, pattern color=red
        },error bars/.cd, y dir=both,y explicit] coordinates {({Autonomous steering}, 0.5255) +- (0.1177,0.0954)({Dynamic Environment},0.4982) +- (0.0678,0.0163)
            }; 
            \addplot[
        brown,
        fill=white,
        postaction={
            pattern=north east lines, pattern color=brown
        },error bars/.cd, y dir=both,y explicit] coordinates {({Autonomous steering}, 0.5586) +- (0.1177, 0.1111)({Dynamic Environment},0.5744) +- (0.0418,0.0261)
            }; 

        \end{axis}
    \end{tikzpicture}}
    
    \caption{  Bias \& Coverage ($S_1$). The controllers are biased towards different input data distributions and cover the full context's space.  
    }
    \label{fig:setting1}
\end{subfigure}
\begin{subfigure}[b]{0.49\textwidth}
\centering
    \scalebox{0.65}{\begin{tikzpicture}
    \begin{axis}[
            symbolic x coords={{Autonomous steering}, {Dynamic Environment},  
            },
            enlarge x limits={abs=0.3\linewidth},
            width=1.5\linewidth,
            height=0.6\linewidth,
            xtick=data,
            ylabel={Average reward},
            ymin=0.0, ymax=1,
            ytick={0.1,0.2,0.3,0.4,0.5,0.6,0.7,0.8,0.9,1.0},
            ybar,
            bar width=6pt,
            legend image code/.code={
                \draw [#1] (0cm,-0.1cm) rectangle (0.2cm,0.25cm); },
            legend style={at={(0.5,-0.2)},anchor=north,legend columns=-1},
        ]
        \addplot[pattern=north east lines] coordinates {({Autonomous steering},0.3649635)({Dynamic Environment},0.124)
        }; 
        \addplot coordinates {({Autonomous steering},0)({Dynamic Environment},0)
        }; 
        \addplot[pattern=horizontal lines] coordinates {({Autonomous steering},0.428)({Dynamic Environment},0.4177897574123988)
        }; 
        \addplot coordinates {({Autonomous steering},0)({Dynamic Environment},0)
        }; 
        \addplot+[error bars/.cd, y dir=both,y explicit] coordinates {({Dynamic Environment},0.4207) +- (0.0213,0.0227)({Autonomous steering},0.3858533398011414) +- (0.055,0.05)
        }; 
        \addplot+[error bars/.cd, y dir=both,y explicit] coordinates {({Dynamic Environment},0.4362) +- (0.0106,0.0344)({Autonomous steering},0.4330989813202281) +- (0.04,0.03)
        }; 
        \addplot+[error bars/.cd, y dir=both,y explicit] coordinates {({Dynamic Environment},0.4572) +- (0.0321,0.0362)({Autonomous steering},0.4777137994926482) +- (0.013,0.032)
        }; 
        \addplot+[error bars/.cd, y dir=both,y explicit] coordinates {({Dynamic Environment},0.5062) +- (0.0579,0.0424)({Autonomous steering}, 0.5475795606661843) +- (0.0276,0.0083)
        }; 
        \addplot+[error bars/.cd, y dir=both,y explicit] coordinates {({Dynamic Environment},0.5553) +- (0.0539,0.0380)({Autonomous steering}, 0.6186666666666667) +- (0.018, 0.02213)
        }; 
        \addplot+[
    magenta,
    fill=white,
    postaction={
        pattern=north east lines, pattern color=magenta
    },error bars/.cd, y dir=both,y explicit] coordinates {({Autonomous steering},0.409) +- (0.03,0.03)({Dynamic Environment},0.3273) +- (0.0427,0.0269)
        }; 
        \addplot[
    green,
    fill=white,
    postaction={
        pattern=north east lines, pattern color=green
    },error bars/.cd, y dir=both,y explicit] coordinates {({Autonomous steering},0.439) +- (0.045,0.04)({Dynamic Environment},0.3388) +- (0.0373,0.0272)
        }; 
        \addplot[
    blue,
    fill=white,
    postaction={
        pattern=north east lines, pattern color=blue
    },error bars/.cd, y dir=both,y explicit] coordinates {({Autonomous steering},0.449) +- (0.046,0.047)({Dynamic Environment},0.3465) +- (0.0375,0.0247)
        }; 
        \addplot[
    red,
    fill=white,
    postaction={
        pattern=north east lines, pattern color=red
    },error bars/.cd, y dir=both,y explicit] coordinates {({Autonomous steering}, 0.487) +- (0.052,0.02)({Dynamic Environment},0.3596) +- (0.0415,0.0229)
        }; 
        \addplot[
    brown,
    fill=white,
    postaction={
        pattern=north east lines, pattern color=brown
    },error bars/.cd, y dir=both,y explicit] coordinates {({Autonomous steering}, 0.515) +- (0.035, 0.02)({Dynamic Environment},0.3728) +- (0.0460,0.0271)
        }; 

        \end{axis}
    \end{tikzpicture}}

    \caption{Bias \& No Coverage ($S_2$). The controllers are biased towards different input data distributions, and do not cover the full context's space.
    }
    \label{fig:setting2}
    
\end{subfigure}

\captionsetup[subfigure]{oneside,margin={-0.05\linewidth,-0.1\linewidth}}
\begin{subfigure}[b]{0.95\textwidth}
\centering

   \scalebox{0.75}{
    \begin{tikzpicture}
    \begin{axis}[
            symbolic x coords={{Autonomous steering}, {Dynamic Environment}, {Dynamic Environment improved},    
            },
            enlarge x limits={abs=0.15\linewidth},
            width=1.1\linewidth,
            height=0.27\linewidth,
            xtick=data,
            ylabel={Average reward},
            ymin=0.0, ymax=1,
            ytick={0.1,0.2,0.3,0.4,0.5,0.6,0.7,0.8,0.9,1.0},
            ybar,
            bar width=7pt,
            legend image code/.code={
                \draw [#1] (0cm,-0.1cm) rectangle (0.2cm,0.25cm); },
            legend style={at={(0.5,-0.2)},anchor=north,legend columns=-1},
        ]
        \addplot[pattern=north east lines] coordinates {({Dynamic Environment},0.58)({Autonomous steering}, 0.82997118)({Dynamic Environment improved}, 0.63829787)
        }; 
        \addplot coordinates {({Autonomous steering},0)
        }; 
        \addplot[pattern=horizontal lines] coordinates {({Dynamic Environment},0.628)({Dynamic Environment improved}, 0.6677740863787376)({Autonomous steering}, 0.592)
        }; 
        \addplot coordinates {({Autonomous steering},0)
        }; 
        \addplot+[error bars/.cd, y dir=both,y explicit] coordinates {({Dynamic Environment},0.6707) +- (0.0273,0.0347)({Autonomous steering}, 0.828666666) +- (0.006,0.004)({Dynamic Environment improved}, 0.7433) +- (0.0347,0.0313)
        }; 
        \addplot+[error bars/.cd, y dir=both,y explicit] coordinates {({Dynamic Environment},0.7065) +- (0.0603,0.0503)({Autonomous steering}, 0.8333334) +- (0.011, 0.009)({Dynamic Environment improved}, 0.8181) +- (0.0472,0.0594)
        }; 
        \addplot+[error bars/.cd, y dir=both,y explicit] coordinates {({Dynamic Environment},0.7687) +- (0.0663,0.0790)({Autonomous steering}, 0.8362809933)+-(0.006,0.008)({Dynamic Environment improved}, 0.8532) +- (0.0334,0.0408)
        }; 
        \addplot+[error bars/.cd, y dir=both,y explicit] coordinates {({Dynamic Environment}, 0.8528) +- (0.0460,0.0659)({Autonomous steering}, 0.849733333)+-(0.006,0.0043)({Dynamic Environment improved}, 0.8812) +- (0.0376,0.0499)
        }; 
        \addplot+[error bars/.cd, y dir=both,y explicit] coordinates {({Dynamic Environment}, 0.8989) +- (0.0303, 0.0544)({Autonomous steering}, 0.8665454533)+-(0.004,0.007)({Dynamic Environment improved}, 0.9258) +- (0.0172,0.0457)
        }; 
        \addplot coordinates {({Dynamic Environment},0)
        }; 
        \addplot[
    magenta,
    fill=white,
    postaction={
        pattern=north east lines, pattern color=magenta
    },error bars/.cd, y dir=both,y explicit] coordinates {({Autonomous steering},0.8442) +- (0.0242,0.0124)({Dynamic Environment},0.4559) +- (0.0458,0.0459)({Dynamic Environment improved}, 0.5409) +- (0.0386,0.0207)
        }; 
        \addplot[
    green,
    fill=white,
    postaction={
        pattern=north east lines, pattern color=green
    },error bars/.cd, y dir=both,y explicit] coordinates {({Autonomous steering},0.8539) +- (0.0255,0.0149)({Dynamic Environment},0.4610) +- (0.0864,0.0495)({Dynamic Environment improved}, 0.5431) +- (0.0388,0.0197)
        }; 
        \addplot[
    blue,
    fill=white,
    postaction={
        pattern=north east lines, pattern color=blue
    },error bars/.cd, y dir=both,y explicit] coordinates {({Autonomous steering},0.8615) +- (0.0282,0.0154)({Dynamic Environment},0.4856) +- (0.0972,0.0675)({Dynamic Environment improved}, 0.5452) +- (0.0409,0.0202)
        }; 
        \addplot[
    red,
    fill=white,
    postaction={
        pattern=north east lines, pattern color=red
    },error bars/.cd, y dir=both,y explicit] coordinates {({Autonomous steering}, 0.8772) +- (0.0222,0.0176)({Dynamic Environment},0.5362) +- (0.0722,0.0776)({Dynamic Environment improved}, 0.5589) +- (0.0466,0.0236)
        }; 
        \addplot[
    brown,
    fill=white,
    postaction={
        pattern=north east lines, pattern color=brown
    },error bars/.cd, y dir=both,y explicit] coordinates {({Autonomous steering}, 0.8918) +- (0.0240, 0.0237)({Dynamic Environment},0.5768) +- (0.0479,0.0912)({Dynamic Environment improved}, 0.5812) +- (0.0371,0.0289)
        }; 

        \legend{W. Avg,, MoE,, LR-NSC, LR-FP$5$, LR-FP$10$, LR-FP$20$, LR-FP$30$,, NN-NSC, NN-FP$5$, NN-FP$10$, NN-FP$20$, NN-FP$30$}

        \end{axis}
    \end{tikzpicture}
}
    \caption{No Bias \& Coverage ($S_3$). The controllers are not biased towards different input data distributions and cover the full context's space. 
    }
    \label{fig:setting3}
\end{subfigure}

\caption{Results for RQ2: Comparing to other baselines. Weighted Average and MoE vs two types of monitors: Logistic Regression (LR, filled colored bars) and Neural Network (NN, striped colored bars). The legend key NSC stands for No  Safe Controller, FP represents the tolerance rate of False Positive in Percentage, e.g., FP5 stands for a 5\% tolerance of FP. }

\end{figure*}

In this experiment, we compare our contextual approach to other ensemble-based techniques commonly used in ensemble-based control. 
In regression tasks, like in our scenarios, common ensemble techniques are based on (weighted) averaging methods. This could be simple approaches based on bagging, or one that allows for contextual information to be used in order to dynamically compute the weights, such as mixture-of-experts methods \cite{learningsituationaldriving}. Our comparison is performed over all three settings described in \Cref{fig:biascover}. We especially show that, depending on the setting, our contextual approach has a significant impact on common ensemble approaches, especially for scenarios with more complex dynamics, such as Scenario~2. 
We specifically perform the comparison for two types of monitors, one based on logistic regression (LR) as described in the paper, and one where the monitor is implemented by a neural network (NN). The goal here is to specifically show that logistic regressions, in addition to their value in providing theoretical bounds on the regret,  indeed provide better generalizations compared to neural networks provided with the same amount of data. 
We trained three monitors per scenario and model type (LR, NN). The hyperparameters chosen were $T=1000$, $e=25$, and $n=300$ in each case, and each method is evaluated over $500$ simulations.

Independent of the scenario and the setting, we expect the monitors to switch to a safe 
controller when they do not trust the best controller for a given context. 
Specifically, we define a confidence threshold $1-\varepsilon$ such that, if a monitor returns a controller $c$ for a context $\xi$, and the monitor estimates that the confidence of $c$ satisfying the specifications is less than $1-\varepsilon$, then the monitor will switch to the safe controller. Increasing the confidence threshold will lead to safer systems, but also to an increase in false positives, that is, cases where the monitor unnecessarily switches to the safe controller.

Our results are given in \Cref{fig:setting1,fig:setting2,fig:setting3}. We first describe the setups for our baselines and then present our findings.

\textit{Bagging} The bagging method used computes the weighted average of the output. To estimate the weights, we generated sets of data with $\sim$450K instances, computed the MSE of each controller over that dataset, and then defined the weights as the inverse of the MSE values.

\textit{Mixture of experts.} We compare to an MoE approach where a neural network with a gate layer determines the contribution of each controller  based on contextual input \cite{learningsituationaldriving}. We use the same datasets from Bagging to train the MoE models. \\

\textit{Setting $S_1$: Bias \& Coverage.} In setting $S_1$,  the controllers are biased towards different input data distributions and cover the full context's space. For each scenario, we employ 15 controllers trained on 15 biased datasets generated from different contexts.
The monitors are tested over 500 simulations generated based on the 15 contexts chosen to train the biased controllers. Note that, during simulation, the context can change as explained in \Cref{section:scenarios}, and the monitors may switch to another controller. Specifically, the context variable $d$ can change for Scenario 1, the variables $d_c$ and $d_p$ can change for Scenario 2, and the variable $i$ can change for both at each time step.

The results of both scenarios for LR and the averaging ensemble align with our expectations (\Cref{fig:setting1}). Because the controllers are highly biased towards a different context each, the averaging ensemble does not perform well, violating the safety specification in more than half of the simulations for Scenario 1, and in almost 90\% of them for Scenario 2. However, our monitors learn to assign the best controller to each context, improving the reward by around 30\% in comparison to the averaging ensemble, even when the safety controller is not used. Tolerating rates of False Positives of 5\%, 10\%, 20\%, and 30\%, we can observe that the reward increases even further, to 80\% in Scenario 1, and 70\% in Scenario 2.
For the NN-based monitors, the results show lower performance than for LR-based monitors. This may indicate that NNs require more data to achieve similar performance.

\textit{Setting $S_2$: Bias \& No Coverage}  In $S_2$, the controllers are also biased, but do not cover the full context's space, which allows us to introduce out-of-distribution input data. We employ the same sets of 15 controllers of setting $S_1$, but we train and evaluate the monitors on the full context's spaces defined for each scenario in \Cref{section:scenarios}. 
Because the controllers do not cover the context space, and there are many out-of-distribution data instances, all approaches violate the safety specification for more than half of the simulations (\Cref{fig:setting2}). {Similar to setting $S_1$, because the controllers are highly biased towards a different context each, the averaging ensembles do not perform well.}
It is worth noting that, without using the safe controller, MoE is slightly better than LR for Scenario 1 and as good as LR for Scenario 2. However, because we have access to the confidence of the monitors, we can increase the confidence threshold to use the safe controller. With just 5\% of false positives, the LR-based monitors already perform better than MoE, and we could raise the reward by up to 30\% for Scenario 1 and 20\% for Scenario 2 when the rate of false positives grows to 30\%. The results also show lower performance for NN-based monitors than for LR-based monitors, as observed in $S_1$.

\textit{Setting $S_3$: No Bias \& Coverage} The controllers are not biased, and they cover the full context's space. For each scenario, we employ 15 generic controllers trained on data collected from i.i.d. sampled simulations from all contexts defined in \Cref{section:scenarios}.
For Scenario 1, because the controllers are well-trained on datasets that cover the full context's space, the safety specification satisfaction is high for the LR and NN monitors (\Cref{fig:setting3}). Even if the monitor does not choose the best controller, the performance is still high. In addition, because the controllers are ``not" biased, the reward is also high for the averaging ensemble. For these reasons, our monitors do not find significant differences between the controllers, so the performance of our monitors is slightly worse than the averaging ensemble when the safe controller is not used (FP 0\%). However, if we tolerate a rate of false positives of 10\% or higher, the monitors perform slightly better than the averaging ensemble. The latter indicates that our approach can exploit even the smallest biases in controllers, although if the context's space is more complex, this may require additional rounds of our algorithm.
Surprisingly, MoE performs much worse than the rest. This may be due to overfitting or requiring fine-tuning. 

For Scenario 2, we noticed that the results obtained were worse than for Scenario 1 (\Cref{fig:setting3}). Our main hypothesis was that, because the dynamic urban environment task in Scenario ~2 was much more complex than the one in Scenario~1, the controllers required more training to achieve similar performance. To test this hypothesis, we trained another set of 15 controllers over datasets of $\sim$150K instances each. Certainly, we can observe that all methods perform better on the new set of controllers (Dynamic Environment improved, \Cref{fig:setting3}). We can also observe that, although the controllers are trained on data collected from i.i.d. sampled simulations from all contexts, the LR monitor achieves better performance than the averaging ensemble and MoE, which implies that they can exploit some biases that appeared during training. {The NN monitors perform the worst, which may indicate that they require more simulations to achieve high performance.}

\textit{Conclusion} We have shown the efficacy of our monitors in situations where there is a contextual bias ($S_1$, and $S_3$: Scenario 2). Furthermore, when the controllers are not trained over the whole OD, and OoD data is introduced ($S_2$), the controller chosen by the monitor will have a low confidence value, which could prompt the monitor to switch to the fail-safe to reduce risks. Moreover, when ``no" bias is present, and the controllers cover the whole context's space ($S_3$: Scenario 1), we observed that the non-contextual ensemble and our approach are almost equivalent if the fail-safe is not used. However, our monitors still have access to the safe controller, and the users can now adapt the confidence threshold to make the monitors more or less conservative. Finally, although contextual approaches like MoE can obtain similar or better performance than our monitors in certain situations ($S_2$), fine-tuning may be required, and no statistical guarantees are provided. Lastly, we also experimented with boosting settings.  We refer the reader to this case in Appendix \Cref{section:boosting}. Boosting resulted in even worse performance. Similarly, using NN-based monitors could require more simulations to achieve high performance, and they don't offer statistical guarantees.

\subsection{RQ3: Active bandits vs passive learning} \label{section:exp3}

\begin{figure*}[t]
    \begin{subfigure}[b]{0.31\textwidth}
        \centering
        \scalebox{0.65}{\begin{tikzpicture}
            \begin{axis}[
                width=1.35\linewidth,
                height= \textwidth,
                xlabel={Confidence $1-\varepsilon$},
                ylabel={}, 
                xmin=-0.05, xmax=1.05, ymin=0, ymax=1,
                xtick={0.0,0.2,0.4,0.60,0.8,1.0},
                ytick={0,0.1,0.2,0.3,0.4,0.5,0.6,0.7,0.8,0.9,1.0},
                ymajorgrids=false,
            ]
            \addplot[color=green, mark=x] coordinates {(0.0,0.828)(0.05,0.828)(0.1,0.828)(0.15,0.828)(0.2,0.828)(0.25,0.828)(0.3,0.828)(0.35,0.828)(0.4,0.828)(0.45,0.828)(0.5,0.828)(0.55,0.828)(0.6,0.828)(0.65,0.834)(0.7,0.844)(0.75,0.8486666666666666)(0.8,0.8653333333333334)(0.85,0.8873333333333333)(0.9,0.926)(0.95,0.9766666666666666)(1.0,1.0)};
            \addplot[color=red, mark=o] coordinates {(0.0,0.8442164673025857)(0.05,0.8442164673025857)(0.1,0.8442164673025857)(0.15,0.8442164673025857)(0.2,0.8442164673025857)(0.25,0.8442164673025857)(0.3,0.8442164673025857)(0.35,0.8442164673025857)(0.4,0.8442164673025857)(0.45,0.8442164673025857)(0.5,0.8442164673025857)(0.55,0.8442164673025857)(0.6,0.8442164673025857)(0.65,0.8481151027801684)(0.7,0.8510390793883555)(0.75,0.8587912794131446)(0.8,0.8890044579338149)(0.85,0.9210629021670642)(0.9,0.9482931229837116)(0.95,0.9892296430711371)(1.0,1.0)};
            \addplot[color=blue, mark=triangle] coordinates {(0.0,0.8342136757889215)(0.05,0.8342136757889215)(0.1,0.8342136757889215)(0.15,0.8342136757889215)(0.2,0.8342136757889215)(0.25,0.8342136757889215)(0.3,0.8342136757889215)(0.35,0.8342136757889215)(0.4,0.8342136757889215)(0.45,0.8342136757889215)(0.5,0.8342136757889215)(0.55,0.8342136757889215)(0.6,0.8342136757889215)(0.65,0.8342136757889215)(0.7,0.8342136757889215)(0.75,0.8342136757889215)(0.8,0.8342136757889215)(0.85,0.8342136757889215)(0.9,0.8342136757889215)(0.95,0.8431322934031913)(1.0,1.0)};
            \end{axis} 
            \begin{axis}[
                width=1.35\linewidth,
                height= \textwidth,
                xmin=-0.05, xmax=1.05, ymin=0, ymax=1,
                axis y line*=right, axis x line=none,
                ytick={0,0.1,0.2,0.3,0.4,0.5,0.6,0.7,0.8,0.9,1.0},
                ylabel={}, 
            ]
            \addplot[color=green, mark=x, style=dashed,mark options={solid}] coordinates {(0.0,0.0)(0.05,0.0)(0.1,0.0)(0.15,0.0)(0.2,0.0)(0.25,0.0)(0.3,0.0)(0.35,0.0)(0.4,0.0)(0.45,0.0)(0.5,0.0)(0.55,0.0)(0.6,0.0)(0.65,0.037476184599069325)(0.7,0.05137370523755977)(0.75,0.08238709397255514)(0.8,0.20719679931982116)(0.85,0.348426355100884)(0.9,0.5635170913213204)(0.95,0.8911506358392071)(1.0,1.0)};
            \addplot[color=red, mark=o, style=dashed,mark options={solid}] coordinates {(0.0,0.0)(0.05,0.0)(0.1,0.0)(0.15,0.0)(0.2,0.0)(0.25,0.0)(0.3,0.0)(0.35,0.0)(0.4,0.0)(0.45,0.0)(0.5,0.0)(0.55,0.0)(0.6,0.0)(0.65,0.017667844522968195)(0.7,0.030624263839811542)(0.75,0.0829262800526571)(0.8,0.2752641257441745)(0.85,0.5620020739164291)(0.9,0.7369582662451616)(0.95,0.9277406746148698)(1.0,1.0)};
            \addplot[color=blue, mark=triangle, style=dashed,mark options={solid}] coordinates {(0.0,0.0)(0.05,0.0)(0.1,0.0)(0.15,0.0)(0.2,0.0)(0.25,0.0)(0.3,0.0)(0.35,0.0)(0.4,0.0)(0.45,0.0)(0.5,0.0)(0.55,0.0)(0.6,0.0)(0.65,0.0)(0.7,0.0)(0.75,0.0)(0.8,0.0)(0.85,0.0)(0.9,0.0)(0.95,0.11296884712433448)(1.0,1.0)};
            \end{axis}
        \end{tikzpicture}}
        \caption{Scenario 1.} 
        \label{fig:e3_senario1}
    \end{subfigure}
    \hfill
    \begin{subfigure}[b]{0.31\textwidth}
        \centering
        \scalebox{0.65}{\begin{tikzpicture}
            \begin{axis}[
                width=1.35\linewidth,
                height= \textwidth,
                xlabel={Confidence $1-\varepsilon$},
                ylabel={}, 
                xmin=-0.05, xmax=1.05, ymin=0, ymax=1,
                xtick={0,0.2,0.4,0.6,0.8,1.0},
                ytick={0,0.1,0.2,0.3,0.4,0.5,0.6,0.7,0.8,0.9,1.0},
                ymajorgrids=false 
            ]
            \addplot[color=green, mark=x] coordinates {(0.0,0.41533333333333333)(0.05,0.41533333333333333)(0.1,0.4266666666666667)(0.15,0.43133333333333335)(0.2,0.43599999999999994)(0.25,0.45999999999999996)(0.3,0.47866666666666663)(0.35,0.516)(0.4,0.5853333333333334)(0.45,0.6586666666666666)(0.5,0.742)(0.55,0.8046666666666668)(0.6,0.8306666666666667)(0.65,0.8706666666666667)(0.7,0.8933333333333334)(0.75,0.9199999999999999)(0.8,0.9446666666666667)(0.85,0.9586666666666668)(0.9,0.9806666666666667)(0.95,0.9946666666666667)(1.0,1.0)};
            \addplot[color=red, mark=o] coordinates {(0.0,0.4559140534262485)(0.05,0.4559140534262485)(0.1,0.4559140534262485)(0.15,0.4565807200929152)(0.2,0.4579140534262485)(0.25,0.47421602787456446)(0.3,0.4930336817653891)(0.35,0.5057003484320558)(0.4,0.5376376306620209)(0.45,0.5702624854819976)(0.5,0.6198977932636468)(0.55,0.6689547038327527)(0.6,0.7276678281068524)(0.65,0.7827665505226481)(0.7,0.8147247386759583)(0.75,0.8716933797909409)(0.8,0.9399535423925668)(0.85,0.9712450638792102)(0.9,0.9849012775842044)(0.95,0.9988385598141697)(1.0,1.0)};
            \addplot[color=blue, mark=triangle] coordinates {(0.0,0.6706666666666666)(0.05,0.6706666666666666)(0.1,0.6706666666666666)(0.15,0.6706666666666666)(0.2,0.6706666666666666)(0.25,0.6706666666666666)(0.3,0.6706666666666666)(0.35,0.6706666666666666)(0.4,0.6706666666666666)(0.45,0.6706666666666666)(0.5,0.6706666666666666)(0.55,0.6706666666666666)(0.6,0.6706666666666666)(0.65,0.6706666666666666)(0.7,0.6713333333333332)(0.75,0.6966666666666667)(0.8,0.7426666666666666)(0.85,0.8193333333333334)(0.9,0.886)(0.95,0.9526666666666666)(1.0,1.0)};
            \end{axis}
            \begin{axis}[
                width=1.35\linewidth,
                height= \textwidth,
                xmin=-0.05, xmax=1.05, ymin=0, ymax=1,
                axis y line*=right, axis x line=none,
                ytick={0,0.1,0.2,0.3,0.4,0.5,0.6,0.7,0.8,0.9,1.0},
                ylabel={}, 
            ]
            \addplot[color=green, mark=x, style=dashed,mark options={solid}] coordinates {(0.0,0.0)(0.05,0.018181818181818177)(0.1,0.06315536315536315)(0.15,0.07888407888407888)(0.2,0.1284591951258618)(0.25,0.16286676286676285)(0.3,0.21643418310084975)(0.35,0.345326278659612)(0.4,0.48276414943081614)(0.45,0.6103094436427771)(0.5,0.7482924482924482)(0.55,0.8478755812089145)(0.6,0.8980920314253646)(0.65,0.9302388969055634)(0.7,0.9540484207150873)(0.75,0.9746673080006413)(0.8,0.985185185185185)(0.85,0.9902356902356901)(0.9,0.9959595959595959)(0.95,1.0)(1.0,1.0)};
            \addplot[color=red, mark=o, style=dashed,mark options={solid}] coordinates {(0.0,0.0)(0.05,0.0)(0.1,0.005025125628140704)(0.15,0.015636843435052356)(0.2,0.04139719414773079)(0.25,0.08745456066569263)(0.3,0.10818978575023035)(0.35,0.15487251706432326)(0.4,0.20212125414867219)(0.45,0.2878285644291945)(0.5,0.3800766806879669)(0.55,0.46981397998637114)(0.6,0.5752752757366421)(0.65,0.6712699940403538)(0.7,0.7599808710301557)(0.75,0.8814394242537628)(0.8,0.945525733115395)(0.85,0.972550710415805)(0.9,0.9933498337458436)(0.95,1.0)(1.0,1.0)};
            \addplot[color=blue, mark=triangle, style=dashed,mark options={solid}] coordinates {(0.0,0.0)(0.05,0.0)(0.1,0.0)(0.15,0.0)(0.2,0.0)(0.25,0.0)(0.3,0.0)(0.35,0.0)(0.4,0.0)(0.45,0.0)(0.5,0.0)(0.55,0.0)(0.6,0.0)(0.65,0.003902566788894709)(0.7,0.006876782492287993)(0.75,0.042467350759446965)(0.8,0.07786341651960312)(0.85,0.14311954445049088)(0.9,0.2563402618494239)(0.95,0.48154068362941)(1.0,1.0)};
            \end{axis}
        \end{tikzpicture}}
        \caption{Scenario 2.} 
        \label{fig:e3_senario2}
    \end{subfigure}
    \hfill
    \begin{subfigure}[b]{0.31\textwidth}
        \centering
        \scalebox{0.65}{\begin{tikzpicture}
            \begin{axis}[
                width=1.35\linewidth,
                height= \textwidth,
                xlabel={Confidence $1-\varepsilon$},
                ylabel={}, 
                xmin=-0.05, xmax=1.05, ymin=0, ymax=1,
                xtick={0,0.2,0.4,0.6,0.8,1.0},
                ytick={0,0.1,0.2,0.3,0.4,0.5,0.6,0.7,0.8,0.9,1.0},
                ymajorgrids=false,
            ]
            \addplot[color=green, mark=x] coordinates {(0.0,0.43527697839624446)(0.05,0.43527697839624446)(0.1,0.43527697839624446)(0.15,0.43527697839624446)(0.2,0.43527697839624446)(0.25,0.43527697839624446)(0.3,0.4514025284667486)(0.35,0.4654895095262068)(0.4,0.5058043865383315)(0.45,0.5452747994032398)(0.5,0.5825774064306174)(0.55,0.6802959757088197)(0.6,0.7470773525819397)(0.65,0.8311035806448649)(0.7,0.8754672938159177)(0.75,0.9118598999332944)(0.8,0.9533778148457047)(0.85,0.9756556389583912)(0.9,0.9898248540450375)(0.95,0.9979612640163099)(1.0,1.0)};
            \addplot[color=red, mark=o] coordinates {(0.0,0.538)(0.05,0.538)(0.1,0.538)(0.15,0.538)(0.2,0.538)(0.25,0.538)(0.3,0.538)(0.35,0.538)(0.4,0.538)(0.45,0.538)(0.5,0.538)(0.55,0.5426666666666667)(0.6,0.548)(0.65,0.5573333333333333)(0.7,0.598)(0.75,0.6353333333333334)(0.8,0.7186666666666666)(0.85,0.7866666666666666)(0.9,0.842)(0.95,0.9186666666666666)(1.0,1.0)};
            \addplot[color=blue, mark=triangle] coordinates {(0.0,0.7433333333333333)(0.05,0.7433333333333333)(0.1,0.7433333333333333)(0.15,0.7433333333333333)(0.2,0.7433333333333333)(0.25,0.7433333333333333)(0.3,0.7433333333333333)(0.35,0.7433333333333333)(0.4,0.7433333333333333)(0.45,0.7433333333333333)(0.5,0.7433333333333333)(0.55,0.746)(0.6,0.7466666666666667)(0.65,0.7513333333333333)(0.7,0.7673333333333333)(0.75,0.7846666666666667)(0.8,0.8286666666666666)(0.85,0.8613333333333334)(0.9,0.8846666666666666)(0.95,0.9346666666666666)(1.0,1.0)};
            \end{axis}
            \begin{axis}[
                width=1.35\linewidth,
                height= \textwidth,
                xmin=-0.05, xmax=1.05, ymin=0, ymax=1,
                axis y line*=right, axis x line=none,
                ytick={0,0.1,0.2,0.3,0.4,0.5,0.6,0.7,0.8,0.9,1.0},
            ]
            \addplot[color=green, mark=x, style=dashed,mark options={solid}] coordinates {(0.0,0.0)(0.05,0.0)(0.1,0.008403361344537815)(0.15,0.02647165736523617)(0.2,0.03436183633758847)(0.25,0.07878802761562641)(0.3,0.15256467567943208)(0.35,0.2592648462056579)(0.4,0.44830372511600064)(0.45,0.5783582039686103)(0.5,0.6390182981589582)(0.55,0.7318119280154809)(0.6,0.8373315452135053)(0.65,0.8792312487972288)(0.7,0.9333798404858126)(0.75,0.9714138040712467)(0.8,0.9791666666666666)(0.85,0.9869791666666666)(0.9,0.9895833333333334)(0.95,1.0)(1.0,1.0)};
            \addplot[color=red, mark=o, style=dashed,mark options={solid}] coordinates {(0.0,0.0)(0.05,0.0)(0.1,0.0)(0.15,0.0)(0.2,0.0)(0.25,0.0)(0.3,0.0)(0.35,0.0)(0.4,0.0)(0.45,0.0)(0.5,0.0)(0.55,0.031531531531531536)(0.6,0.05584576172811467)(0.65,0.11461769893142441)(0.7,0.21479266381227166)(0.75,0.31833177715530653)(0.8,0.417501535148594)(0.85,0.5419108183814066)(0.9,0.6664151546504488)(0.95,0.9010377324102814)(1.0,1.0)};
            \addplot[color=blue, mark=triangle, style=dashed,mark options={solid}] coordinates {(0.0,0.0)(0.05,0.0)(0.1,0.0)(0.15,0.0)(0.2,0.0)(0.25,0.0)(0.3,0.0)(0.35,0.0)(0.4,0.0)(0.45,0.0)(0.5,0.0)(0.55,0.0)(0.6,0.0)(0.65,0.0)(0.7,0.0)(0.75,0.031531531531531536)(0.8,0.05584576172811467)(0.85,0.11461769893142441)(0.9,0.21479266381227166)(0.95,0.31833177715530653)(1.0,1.0)};
            \end{axis}
        \end{tikzpicture}}
        \caption{Scenario 2 with improved controllers.} 
        \label{fig:e3_senario2_1}
    \end{subfigure}

    \begin{subfigure}[b]{\textwidth}
    \centering
    \scalebox{0.95}{
    \begin{tikzpicture}
    \begin{axis}[
        hide axis,
        width=0.1\linewidth, height=0.1\linewidth,
        xmin=0, xmax=1, ymin=0, ymax=1,
        legend style={
            draw=none, 
            legend columns=6, 
            at={(0.5,0.5)}, 
            anchor=center, 
            column sep=1pt, 
            font=\small,
            cells={anchor=west} 
        }
    ]

    \addlegendimage{color=green, mark=x, mark options={solid}, line width=1pt} 
    \addlegendentry{Reward Passive NN}
    
    \addlegendimage{color=red, mark=o, mark options={solid}, line width=1pt} 
    \addlegendentry{Reward Active NN}
    
    \addlegendimage{color=blue, mark=triangle, mark options={solid}, line width=1pt} 
    \addlegendentry{Reward Active LR}
    
    \addlegendimage{color=green, mark=x, style=dashed, mark options={solid}, line width=1pt} 
    \addlegendentry{FP Passive NN}
    
    \addlegendimage{color=red, mark=o, style=dashed, mark options={solid}, line width=1pt} 
    \addlegendentry{FP Active NN}
    
    \addlegendimage{color=blue, mark=triangle, style=dashed, mark options={solid}, line width=1pt} 
    \addlegendentry{FP Active LR}
    
    \end{axis}
    \end{tikzpicture}}
    \end{subfigure}

    \caption{Results for RQ3: Comparison of passive vs. active learned monitors across different scenarios.}
    \label{fig:e3_combined}

\end{figure*}
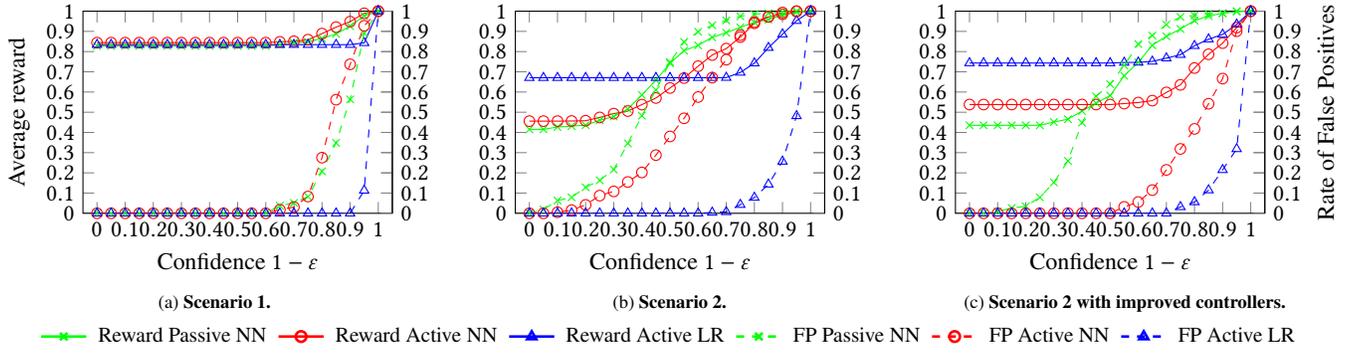

\begin{figure*}[t]
    \centering

    \begin{subfigure}[b]{0.31\textwidth}
        \centering
        \scalebox{0.65}{\begin{tikzpicture}
            \begin{axis}[
                width=1.35\linewidth,
                height= \textwidth,
                xlabel={Confidence $1-\varepsilon$},
                xmin=-0.05, xmax=1.05, ymin=0, ymax=1,
                xtick={0,0.2,0.4,0.6,0.8,1.0},
                ytick={0,0.1,0.2,0.3,0.4,0.5,0.6,0.7,0.8,0.9,1.0},
                ymajorgrids=false,
            ]
            \addplot[color=blue, mark=x] coordinates {(0.0,0.7990314769975787)(0.05,0.7990314769975787)(0.1,0.7990314769975787)(0.15,0.7990314769975787)(0.2,0.7990314769975787)(0.25,0.7990314769975787)(0.3,0.7990314769975787)(0.35,0.7990314769975787)(0.4,0.7990314769975787)(0.45,0.7990314769975787)(0.5,0.7990314769975787)(0.55,0.7990314769975787)(0.6,0.7990314769975787)(0.65,0.7990314769975787)(0.7,0.801452784503632)(0.75,0.8256658595641646)(0.8,0.8547215496368039)(0.85,0.9225181598062954)(0.9,0.9903147699757869)(0.95,1.0)(1.0,1.0)};
            \addplot[color=blue, mark=o] coordinates {(0.0,0.8044280442804428)(0.05,0.8044280442804428)(0.1,0.8044280442804428)(0.15,0.8044280442804428)(0.2,0.8044280442804428)(0.25,0.8044280442804428)(0.3,0.8044280442804428)(0.35,0.8044280442804428)(0.4,0.8044280442804428)(0.45,0.8044280442804428)(0.5,0.8044280442804428)(0.55,0.8044280442804428)(0.6,0.8044280442804428)(0.65,0.8044280442804428)(0.7,0.8044280442804428)(0.75,0.8081180811808117)(0.8,0.8191881918819188)(0.85,0.8487084870848709)(0.9,0.8929889298892989)(0.95,0.996309963099631)(1.0,1.0)};
            \addplot[color=blue, mark=triangle] coordinates {(0.0,0.8342136757889215)(0.05,0.8342136757889215)(0.1,0.8342136757889215)(0.15,0.8342136757889215)(0.2,0.8342136757889215)(0.25,0.8342136757889215)(0.3,0.8342136757889215)(0.35,0.8342136757889215)(0.4,0.8342136757889215)(0.45,0.8342136757889215)(0.5,0.8342136757889215)(0.55,0.8342136757889215)(0.6,0.8342136757889215)(0.65,0.8342136757889215)(0.7,0.8342136757889215)(0.75,0.8342136757889215)(0.8,0.8342136757889215)(0.85,0.8342136757889215)(0.9,0.8342136757889215)(0.95,0.8431322934031913)(1.0,1.0)};
            \end{axis}
            
            \begin{axis}[
                width=1.35\linewidth,
                height= \textwidth,
                xmin=-0.05, xmax=1.05, ymin=0, ymax=1,
                axis y line*=right, axis x line=none,
                ytick={0,0.1,0.2,0.3,0.4,0.5,0.6,0.7,0.8,0.9,1.0},
                ylabel={}, 
            ]
            \addplot[color=orange, mark=x, style=dashed,mark options={solid}] coordinates {(0.0,0.0)(0.05,0.0)(0.1,0.0)(0.15,0.0)(0.2,0.0)(0.25,0.0)(0.3,0.0)(0.35,0.0)(0.4,0.0)(0.45,0.0)(0.5,0.0)(0.55,0.0)(0.6,0.0)(0.65,0.0)(0.7,0.0303030303030303)(0.75,0.21515151515151515)(0.8,0.5484848484848485)(0.85,0.7727272727272727)(0.9,0.9545454545454545)(0.95,1.0)(1.0,1.0)};
            \addplot[color=orange, mark=o, style=dashed,mark options={solid}] coordinates {(0.0,0.0)(0.05,0.0)(0.1,0.0)(0.15,0.0)(0.2,0.0)(0.25,0.0)(0.3,0.0)(0.35,0.0)(0.4,0.0)(0.45,0.0)(0.5,0.0)(0.55,0.0)(0.6,0.0)(0.65,0.0)(0.7,0.0)(0.75,0.013761467889908256)(0.8,0.2477064220183486)(0.85,0.47247706422018354)(0.9,0.5963302752293578)(0.95,0.9954128440366973)(1.0,1.0)};
            \addplot[color=orange, mark=triangle, style=dashed,mark options={solid}] coordinates {(0.0,0.0)(0.05,0.0)(0.1,0.0)(0.15,0.0)(0.2,0.0)(0.25,0.0)(0.3,0.0)(0.35,0.0)(0.4,0.0)(0.45,0.0)(0.5,0.0)(0.55,0.0)(0.6,0.0)(0.65,0.0)(0.7,0.0)(0.75,0.0)(0.8,0.0)(0.85,0.0)(0.9,0.0)(0.95,0.11296884712433448)(1.0,1.0)};
            \end{axis}
        \end{tikzpicture}}
        \caption{Scenario 1.} 
        \label{fig:e4_scenario1}
    \end{subfigure}
    \hfill
    \begin{subfigure}[b]{0.31\textwidth}
        \centering
        \scalebox{0.65}{\begin{tikzpicture}
            \begin{axis}[
                width=1.35\linewidth,
                height= \textwidth,
                xlabel={Confidence $1-\varepsilon$},
                ylabel={}, 
                xmin=-0.05, xmax=1.05, ymin=0, ymax=1,
                xtick={0,0.2,0.4,0.6,0.8,1.0},
                ytick={0,0.1,0.2,0.3,0.4,0.5,0.6,0.7,0.8,0.9,1.0},
                ymajorgrids=false,
            ]
            \addplot[color=blue, mark=x] coordinates {(0.0,0.4982800579058355)(0.05,0.4982800579058355)(0.1,0.4982800579058355)(0.15,0.5003897625471857)(0.2,0.5060203882832567)(0.25,0.5202034672586878)(0.3,0.5392032571841857)(0.35,0.5767091428922188)(0.4,0.6206246456660792)(0.45,0.6494898735320996)(0.5,0.7004252769571148)(0.55,0.7343516799669372)(0.6,0.7683360155005244)(0.65,0.8253674492799705)(0.7,0.8700654319166077)(0.75,0.918074261145863)(0.8,0.9407369916797533)(0.85,0.9654118241825108)(0.9,0.9816306432740513)(0.95,0.9950832653736601)(1.0,1.0)};
            \addplot[color=blue, mark=o] coordinates {(0.0,0.6304184447700408)(0.05,0.6304184447700408)(0.1,0.6304184447700408)(0.15,0.6325281494113911)(0.2,0.6374584950354172)(0.25,0.6502487152476344)(0.3,0.6643347253151975)(0.35,0.6919586073166166)(0.4,0.7301967418245372)(0.45,0.7562354191710711)(0.5,0.8029194446255431)(0.55,0.8318912347843724)(0.6,0.8616289390398871)(0.65,0.907358288354133)(0.7,0.9427976281025545)(0.75,0.9767079991109006)(0.8,0.9887582690623548)(0.85,0.9978932501270551)(0.9,1.0)(0.95,1.0)(1.0,1.0)};
            \addplot[color=blue, mark=triangle] coordinates {(0.0,0.6706666666666666)(0.05,0.6706666666666666)(0.1,0.6706666666666666)(0.15,0.6706666666666666)(0.2,0.6706666666666666)(0.25,0.6706666666666666)(0.3,0.6706666666666666)(0.35,0.6706666666666666)(0.4,0.6706666666666666)(0.45,0.6706666666666666)(0.5,0.6706666666666666)(0.55,0.6706666666666666)(0.6,0.6706666666666666)(0.65,0.6706666666666666)(0.7,0.6713333333333332)(0.75,0.6966666666666667)(0.8,0.7426666666666666)(0.85,0.8193333333333334)(0.9,0.886)(0.95,0.9526666666666666)(1.0,1.0)};
            \end{axis}
            
            \begin{axis}[
                width=1.35\linewidth,
                height= \textwidth,
                xmin=-0.05, xmax=1.05, ymin=0, ymax=1,
                axis y line*=right, axis x line=none,
                ytick={0,0.1,0.2,0.3,0.4,0.5,0.6,0.7,0.8,0.9,1.0},
                ylabel={}, 
            ]
            \addplot[color=orange, mark=x, style=dashed,mark options={solid}] coordinates {(0.0,0.0)(0.05,0.0)(0.1,0.004545454545454545)(0.15,0.024512827775767018)(0.2,0.0707909106645475)(0.25,0.10928158161151064)(0.3,0.1261475758100304)(0.35,0.18780587596582057)(0.4,0.26035454073363024)(0.45,0.35289995331539387)(0.5,0.45141942184524836)(0.55,0.5406470801139314)(0.6,0.6403323524320581)(0.65,0.7220044481512371)(0.7,0.8066326236224106)(0.75,0.8649485157967068)(0.8,0.8986356554990794)(0.85,0.9446727094381586)(0.9,0.9739303192912332)(0.95,0.9953973699256718)(1.0,1.0)};
            \addplot[color=orange, mark=o, style=dashed,mark options={solid}] coordinates {(0.0,0.0)(0.05,0.0)(0.1,0.0)(0.15,0.009530859628018876)(0.2,0.03839302873493675)(0.25,0.04471342173757275)(0.3,0.060474810960607195)(0.35,0.14971658920025718)(0.4,0.21093166110284645)(0.45,0.29394290974748083)(0.5,0.3711729458352018)(0.55,0.4484083203431774)(0.6,0.48972621021987167)(0.65,0.5567118178388651)(0.7,0.6238718138528446)(0.75,0.7105481845812188)(0.8,0.7716573778154235)(0.85,0.8523888244324073)(0.9,0.9415300624239276)(0.95,0.9935844051694889)(1.0,1.0)};
            \addplot[color=orange, mark=triangle, style=dashed,mark options={solid}] coordinates {(0.0,0.0)(0.05,0.0)(0.1,0.0)(0.15,0.0)(0.2,0.0)(0.25,0.0)(0.3,0.0)(0.35,0.0)(0.4,0.0)(0.45,0.0)(0.5,0.0)(0.55,0.0)(0.6,0.0)(0.65,0.003902566788894709)(0.7,0.006876782492287993)(0.75,0.042467350759446965)(0.8,0.07786341651960312)(0.85,0.14311954445049088)(0.9,0.2563402618494239)(0.95,0.48154068362941)(1.0,1.0)};
            \end{axis}
        \end{tikzpicture}}
        \caption{Scenario 2.} 
        \label{fig:e4_scenario2}
    \end{subfigure}
    \hfill
    \begin{subfigure}[b]{0.31\textwidth}
        \centering
        \scalebox{0.65}{\begin{tikzpicture}
            \begin{axis}[
                width=1.35\linewidth,
                height= \textwidth,
                xlabel={Confidence $1-\varepsilon$},
                ylabel={}, 
                xmin=-0.05, xmax=1.05, ymin=0, ymax=1,
                xtick={0,0.2,0.4,0.6,0.8,1.0},
                ytick={0,0.1,0.2,0.3,0.4,0.5,0.6,0.7,0.8,0.9,1.0},
                ymajorgrids=false,
            ]
            \addplot[color=blue, mark=x] coordinates {(0.0,0.5925506891796911)(0.05,0.5925506891796911)(0.1,0.5925506891796911)(0.15,0.5934923087653784)(0.2,0.6091463722160521)(0.25,0.6298446193549772)(0.3,0.6636431288597013)(0.35,0.7034622627842966)(0.4,0.76775561886673)(0.45,0.8193799260654252)(0.5,0.8727225951143089)(0.55,0.8948792527135278)(0.6,0.9389848214141999)(0.65,0.9680133704615814)(0.7,0.9884485006518905)(0.75,0.9990583804143126)(0.8,1.0)(0.85,1.0)(0.9,1.0)(0.95,1.0)(1.0,1.0)};
            \addplot[color=blue, mark=o] coordinates {(0.0,0.6304184447700408)(0.05,0.6304184447700408)(0.1,0.6304184447700408)(0.15,0.6325281494113911)(0.2,0.6374584950354172)(0.25,0.6502487152476344)(0.3,0.6643347253151975)(0.35,0.6919586073166166)(0.4,0.7301967418245372)(0.45,0.7562354191710711)(0.5,0.8029194446255431)(0.55,0.8318912347843724)(0.6,0.8616289390398871)(0.65,0.907358288354133)(0.7,0.9427976281025545)(0.75,0.9767079991109006)(0.8,0.9887582690623548)(0.85,0.9978932501270551)(0.9,1.0)(0.95,1.0)(1.0,1.0)};
            \addplot[color=blue, mark=triangle] coordinates {(0.0,0.7433333333333333)(0.05,0.7433333333333333)(0.1,0.7433333333333333)(0.15,0.7433333333333333)(0.2,0.7433333333333333)(0.25,0.7433333333333333)(0.3,0.7433333333333333)(0.35,0.7433333333333333)(0.4,0.7433333333333333)(0.45,0.7433333333333333)(0.5,0.7433333333333333)(0.55,0.746)(0.6,0.7466666666666667)(0.65,0.7513333333333333)(0.7,0.7673333333333333)(0.75,0.7846666666666667)(0.8,0.8286666666666666)(0.85,0.8613333333333334)(0.9,0.8846666666666666)(0.95,0.9346666666666666)(1.0,1.0)};
            \end{axis}
            
            \begin{axis}[
                width=1.35\linewidth,
                height= \textwidth,
                xmin=-0.05, xmax=1.05, ymin=0, ymax=1,
                axis y line*=right, axis x line=none,
                ytick={0,0.1,0.2,0.3,0.4,0.5,0.6,0.7,0.8,0.9,1.0},
            ]
            \addplot[color=orange, mark=x, style=dashed,mark options={solid}] coordinates {(0.0,0.0)(0.05,0.0)(0.1,0.0)(0.15,0.002207505518763797)(0.2,0.03418237391747326)(0.25,0.061957362281341515)(0.3,0.11686768771432167)(0.35,0.17897225775161837)(0.4,0.3044909971157053)(0.45,0.42868144608103637)(0.5,0.49781423969331917)(0.55,0.5700828639177299)(0.6,0.6434892025248878)(0.65,0.7470195010573303)(0.7,0.8272879200460314)(0.75,0.8941886531528724)(0.8,0.9256220902035377)(0.85,0.980200373577857)(0.9,0.9952283919171337)(0.95,1.0)(1.0,1.0)};
            \addplot[color=orange, mark=o, style=dashed,mark options={solid}] coordinates {(0.0,0.0)(0.05,0.0)(0.1,0.0)(0.15,0.0)(0.2,0.0)(0.25,0.0)(0.3,0.007616438356164383)(0.35,0.023232876712328765)(0.4,0.032365296803652965)(0.45,0.09690464679022293)(0.5,0.17122249082281313)(0.55,0.2870015220700152)(0.6,0.3698576416868116)(0.65,0.4454991494314621)(0.7,0.5387787626466111)(0.75,0.6382279523681618)(0.8,0.7006063210672396)(0.85,0.7999681260632107)(0.9,0.918258035634345)(0.95,0.9973333333333333)(1.0,1.0)};
            \addplot[color=orange, mark=triangle, style=dashed,mark options={solid}] coordinates {(0.0,0.0)(0.05,0.0)(0.1,0.0)(0.15,0.0)(0.2,0.0)(0.25,0.0)(0.3,0.0)(0.35,0.0)(0.4,0.0)(0.45,0.0)(0.5,0.0)(0.55,0.0)(0.6,0.0)(0.65,0.0)(0.7,0.0)(0.75,0.031531531531531536)(0.8,0.05584576172811467)(0.85,0.11461769893142441)(0.9,0.21479266381227166)(0.95,0.31833177715530653)(1.0,1.0)};
            \end{axis}
        \end{tikzpicture}}
        \caption{Scenario 2 with improved controllers.} 
        \label{fig:e4_scenario3}
    \end{subfigure}

    \centering
    \scalebox{0.9}{\begin{tikzpicture}
    \begin{axis}[
        hide axis,
        width=0.1\linewidth, height=0.1\linewidth,
        xmin=0, xmax=1, ymin=0, ymax=1,
        legend style={
            draw=none, 
            legend columns=6, 
            at={(0.5,0.5)}, 
            anchor=center, 
            column sep=2pt,
            font=\small,
            cells={anchor=west}
        }
    ]

    \addlegendimage{color=blue, mark=x, mark options={solid}, line width=1pt} 
    \addlegendentry{Reward 1 controller}
    
    \addlegendimage{color=blue, mark=o, mark options={solid}, line width=1pt} 
    \addlegendentry{Reward 5 controllers}
    
    \addlegendimage{color=blue, mark=triangle, mark options={solid}, line width=1pt} 
    \addlegendentry{Reward 15 controllers}
    
    \addlegendimage{color=orange, mark=x, style=dashed, mark options={solid}, line width=1pt} 
    \addlegendentry{FP 1 controller}
    
    \addlegendimage{color=orange, mark=o, style=dashed, mark options={solid}, line width=1pt} 
    \addlegendentry{FP 5 controllers}
    
    \addlegendimage{color=orange, mark=triangle, style=dashed, mark options={solid}, line width=1pt} 
    \addlegendentry{FP 15 controllers}
    
    \end{axis}
    \end{tikzpicture}}
   
    \par \vspace{-0.2cm}
    \caption{Results for RQ4: Simplex vs. multi-Simplex.} 
    \label{fig:exp4_specs}

\end{figure*}
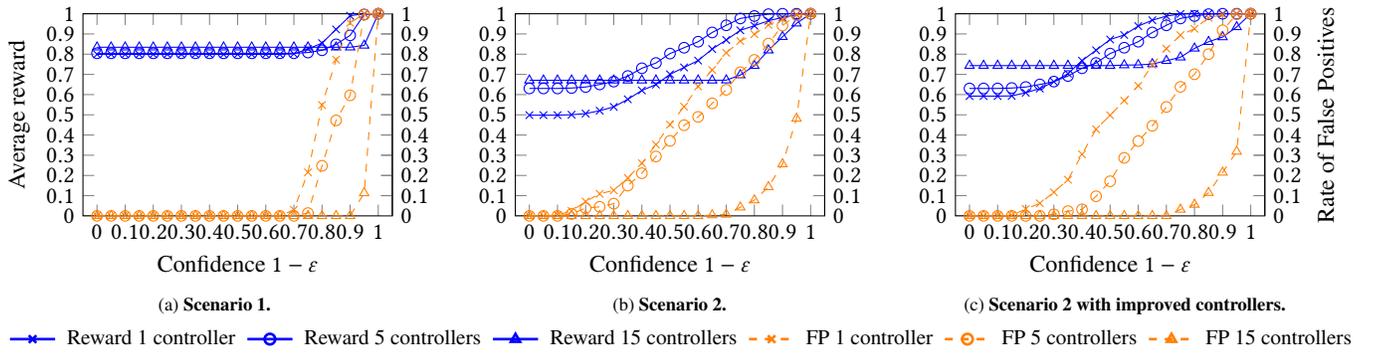

In this section, we compare the efficacy of monitors trained using two distinct data collection strategies: passive learning and active learning. Both approaches are evaluated on the same controllers of Setting~$S_3$ defined in \Cref{section:exp2} for both Scenario 1 and Scenario 2, using the same hyperparameters. Each trained monitor is evaluated over 500 simulations, reporting average reward and the tolerance rate of FP against the confidence threshold $1-\varepsilon$.

\textit{Passive learning (Passive NN).} 
The passive learning approach trains an NN monitor on a dataset generated through random sampling. Specifically, we randomly select (context, controller) pairs and observe the corresponding rewards
to form the training data.

\textit{Active learning (Active NN / Active LR).}
The active learning approach employs our  active query strategy to guide data collection (as discussed in the \emph{Select context and controller} step of \Cref{sec:learning_contextual_monitors}) to learn an LR model. The latter model is denoted by Active LR. An NN trained on this actively collected dataset is denoted Active~NN.

To ensure a fair comparison, Passive and Active NNs use identical architectures, hyperparameters, and training-set sizes.

The comparison results are shown in \Cref{fig:e3_combined}. For Scenario~1, a task with a rather lower complexity (\Cref{fig:e3_senario1}),  the Passive NN and Active NN monitors exhibit nearly identical performance. This suggests that informative, high-uncertainty data points are common enough that a passive approach is sufficient to learn an effective monitor.
However, when compared to Active LR, which gains similar rewards but maintains a much lower tolerance rate of FP, the results indicate that the LR-based monitor performs as well as the complex NN architecture without the need for extra hyperparameter tuning.

The results for Scenario 2 further reveal the impact of the data-sampling strategy (\Cref{fig:e3_senario2}):

\begin{itemize}
    \item \textit{Active LR vs Active NN.} Active LR achieves the highest rewards initially and maintains consistent performance across all confidence levels. Conversely, Active NN starts with relatively low rewards and triggers the safety backup for lower values of the confidence threshold, causing the FP to increase as well. Notably,  when the rewards of Active NN exceed those of the Active LR (when the confidence threshold has been relaxed to nearly 0.55), the Active LR's FP remains at 0\%, while the Active NN's FP already reaches around 47\%. 

\item \textit{Active NN vs Passive NN.} Both NN monitors achieve similar rewards at the outset, with the Active NN performing slightly better. Establishing a confidence threshold of 0.3 leads Passive NN to gain in reward. However, this correlates with a sharp spike in its FP. In contrast, Active NN maintains a lower FP. This suggests that active learning produces a ``smarter" and less conservative monitor that learns a more accurate decision boundary and avoids unnecessary fail-safe activations. 

\end{itemize}

\textit{Validation with improved controllers.} In another experiment, we employed the 15 controllers trained on a larger dataset of \Cref{section:exp2} - Setting $S_3$. The results in Figure~\ref{fig:e3_senario2_1} show that active learning further increases the reward while reducing the FP rate, indicating that actively learned monitors better exploit higher-quality controllers.

\textit{Conclusion.} The active learning approach trains the model to be less conservative, enabling it to make more autonomous decisions rather than solely relying on the safety controller, particularly in situations where it is unable to make a clear judgment based on the current context. Conversely, the passively learned monitor loses basic ``selection" functionality early on and simply relies on the safety fallback to gain reward at the cost of high conservatism.

\subsection{RQ4: Simplex vs multi-Simplex} \label{section:exp4}
\textit{Study the effects of incorporating more controllers.}

This experiment aims to show the relationship between the false positive rates and the number of controllers. We trained LR-based monitors on three sets of controllers taken from $S_3$. The sets have 1, 5, and 15 controllers, respectively. We do this for the three sets of controllers available in \Cref{section:exp2}: the 15 controllers of Scenario 1 (Autonomous steering), and the two sets for Scenario 2 (Dynamic environment). 
We trained three monitors, one per set of controllers. The hyperparameters chosen were $T=1000$, $e=25$, and $n=300$, as in \Cref{section:exp2}. The monitors are evaluated over 500 simulations, generated using the same random seeds, for a fair comparison.
Because the controllers of $S_3$ are trained on diverse enough data, there is only a small difference between regarding their performance, and the reward is similar for the three sets as a consequence. Nevertheless, the reward grows with the number of controllers (see \Cref{fig:e4_scenario1}), and the set with 15 controllers has a reward that is 3\% higher than the case with just one controller. However, it is important to notice that, \textbf{for the same confidence threshold, the false positives rate decreases significantly} with the number of controllers, since the confidence of the monitor on the best controller increases. Similar conclusions can be drawn from \Cref{fig:e4_scenario2,fig:e4_scenario3}.

\subsection{Monitoring computation overhead }

The LR-based monitors select a controller by multiplying the parameters matrix $\theta$ with the context vector $\xi$, and then choosing the controller with the lowest probability of violating the safety specification.
Given a context, an LR-based monitor takes an average of $4.51\mu s$ to select the controller. An NN-based monitor takes an average of $204.61\mu s$ on CPU, although this computation could be accelerated using a GPU. The average was computed over 100 simulations.

\section{CONCLUSION}
We presented a new approach to learning safety monitors for \acps based on contextual learning. Our approach utilizes methods from contextual bandits to learn monitors. Our results show the significant impact of contextualizing monitors in selecting the right control and balancing safety and performance. 
In this paper, we focused on positional contexts based on static features. In the future, we plan to expand to state-based contexts, adapting both theory and algorithms. We also plan to further investigate the selection of feature spaces, as the effectiveness of our approach relies on the choice of the context space. In general, the choice of contexts must be constrained to features that are reliably monitorable at runtime, and should be selected with this consideration in mind.

\newpage
\bibliographystyle{IEEEtran}
\balance
\bibliography{ref}

\begin{thebibliography}{10}
\providecommand{\url}[1]{#1}
\csname url@samestyle\endcsname
\providecommand{\newblock}{\relax}
\providecommand{\bibinfo}[2]{#2}
\providecommand{\BIBentrySTDinterwordspacing}{\spaceskip=0pt\relax}
\providecommand{\BIBentryALTinterwordstretchfactor}{4}
\providecommand{\BIBentryALTinterwordspacing}{\spaceskip=\fontdimen2\font plus
\BIBentryALTinterwordstretchfactor\fontdimen3\font minus \fontdimen4\font\relax}
\providecommand{\BIBforeignlanguage}[2]{{%
\expandafter\ifx\csname l@#1\endcsname\relax
\typeout{** WARNING: IEEEtran.bst: No hyphenation pattern has been}%
\typeout{** loaded for the language `#1'. Using the pattern for}%
\typeout{** the default language instead.}%
\else
\language=\csname l@#1\endcsname
\fi
#2}}
\providecommand{\BIBdecl}{\relax}
\BIBdecl

\bibitem{Bojarski2016EndTE}
M.~Bojarski, D.~W. del Testa, D.~Dworakowski, B.~Firner, B.~Flepp, P.~Goyal, L.~D. Jackel, M.~Monfort, U.~Muller, J.~Zhang, X.~Zhang, J.~Zhao, and K.~Zieba, ``End to end learning for self-driving cars,'' \emph{ArXiv}, vol. abs/1604.07316, 2016.

\bibitem{7778091}
K.~D. Julian, J.~Lopez, J.~S. Brush, M.~P. Owen, and M.~J. Kochenderfer, ``Policy compression for aircraft collision avoidance systems,'' in \emph{2016 IEEE/AIAA 35th Digital Avionics Systems Conference (DASC)}, 2016, pp. 1--10.

\bibitem{amodei2016concreteproblemsaisafety}
\BIBentryALTinterwordspacing
D.~Amodei, C.~Olah, J.~Steinhardt, P.~Christiano, J.~Schulman, and D.~Mané, ``Concrete problems in {AI} safety,'' 2016. [Online]. Available: \url{https://arxiv.org/abs/1606.06565}
\BIBentrySTDinterwordspacing

\bibitem{seshia-arxiv16}
\BIBentryALTinterwordspacing
S.~A. Seshia, D.~Sadigh, and S.~S. Sastry, ``Toward verified artificial intelligence,'' \emph{Commun. ACM}, vol.~65, no.~7, p. 46–55, Jun. 2022. [Online]. Available: \url{https://doi.org/10.1145/3503914}
\BIBentrySTDinterwordspacing

\bibitem{simplex}
L.~Sha, ``\BIBforeignlanguage{English (US)}{Using simplicity to control complexity},'' \emph{\BIBforeignlanguage{English (US)}{IEEE Software}}, vol.~18, no.~4, pp. 20--28, Jul. 2001.

\bibitem{torfah22}
H.~Torfah, C.~Xie, S.~Junges, M.~Vazquez-Chanlatte, and S.~A. Seshia, ``Learning monitorable operational design domains for assured autonomy,'' in \emph{Automated Technology for Verification and Analysis: 20th International Symposium, ATVA 2022, Virtual Event, October 25–28, 2022, Proceedings}.\hskip 1em plus 0.5em minus 0.4em\relax Berlin, Heidelberg: Springer-Verlag, 2022, p. 3–22.

\bibitem{learningsituationaldriving}
E.~Ohn-Bar, A.~Prakash, A.~Behl, K.~Chitta, and A.~Geiger, ``Learning situational driving,'' in \emph{2020 IEEE/CVF Conference on Computer Vision and Pattern Recognition (CVPR)}, 2020, pp. 11\,293--11\,302.

\bibitem{NIPS2007_4b04a686}
J.~Langford and T.~Zhang, ``The epoch-greedy algorithm for multi-armed bandits with side information,'' in \emph{Advances in Neural Information Processing Systems}, J.~Platt, D.~Koller, Y.~Singer, and S.~Roweis, Eds., vol.~20.\hskip 1em plus 0.5em minus 0.4em\relax Curran Associates, Inc., 2007.

\bibitem{lattimore2020bandit}
T.~Lattimore and C.~Szepesv{\'a}ri, \emph{Bandit algorithms}.\hskip 1em plus 0.5em minus 0.4em\relax Cambridge University Press, 2020.

\bibitem{Das2024ActivePO}
N.~Das, S.~Chakraborty, A.~Pacchiano, and S.~R. Chowdhury, ``Active preference optimization for sample efficient {RLHF},'' 2024.

\bibitem{BouneffoufRA20}
D.~Bouneffouf, I.~Rish, and C.~C. Aggarwal, ``Survey on applications of multi-armed and contextual bandits,'' in \emph{{IEEE} Congress on Evolutionary Computation, {CEC} 2020, Glasgow, United Kingdom, July 19-24, 2020}.\hskip 1em plus 0.5em minus 0.4em\relax {IEEE}, 2020, pp. 1--8.

\bibitem{Bartocci2018}
E.~Bartocci, J.~Deshmukh, A.~Donz{\'e}, G.~Fainekos, O.~Maler, D.~Ni{\v{c}}kovi{\'{c}}, and S.~Sankaranarayanan, \emph{Specification-Based Monitoring of Cyber-Physical Systems: A Survey on Theory, Tools and Applications}.\hskip 1em plus 0.5em minus 0.4em\relax Cham: Springer International Publishing, 2018, pp. 135--175.

\bibitem{DBLP:conf/rv/TorfahJFS21}
\BIBentryALTinterwordspacing
H.~Torfah, S.~Junges, D.~J. Fremont, and S.~A. Seshia, ``Formal analysis of ai-based autonomy: From modeling to runtime assurance,'' in \emph{Runtime Verification - 21st International Conference, {RV} 2021, Virtual Event, October 11-14, 2021, Proceedings}, ser. Lecture Notes in Computer Science, L.~Feng and D.~Fisman, Eds., vol. 12974.\hskip 1em plus 0.5em minus 0.4em\relax Springer, 2021, pp. 311--330. [Online]. Available: \url{https://doi.org/10.1007/978-3-030-88494-9\_19}
\BIBentrySTDinterwordspacing

\bibitem{DBLP:journals/fmsd/MitschP16}
S.~Mitsch and A.~Platzer, ``Modelplex: verified runtime validation of verified cyber-physical system models,'' \emph{Formal Methods Syst. Des.}, vol.~49, no. 1-2, pp. 33--74, 2016.

\bibitem{desai-rv17}
A.~Desai, T.~Dreossi, and S.~A. Seshia, ``Combining model checking and runtime verification for safe robotics,'' in \emph{17th International Conference on Runtime Verification (RV)}, 2017, pp. 172--189.

\bibitem{10.1145/3716550.3722021}
\BIBentryALTinterwordspacing
E.~Nesterini, E.~Bartocci, A.~Gambi, D.~Nickovic, S.~A. Seshia, and H.~Torfah, ``Mining specifications for predictive safety monitoring,'' in \emph{Proceedings of the ACM/IEEE 16th International Conference on Cyber-Physical Systems (with CPS-IoT Week 2025)}, ser. ICCPS '25.\hskip 1em plus 0.5em minus 0.4em\relax New York, NY, USA: Association for Computing Machinery, 2025. [Online]. Available: \url{https://doi.org/10.1145/3716550.3722021}
\BIBentrySTDinterwordspacing

\bibitem{10.5555/3504035.3504361}
M.~Alshiekh, R.~Bloem, R.~Ehlers, B.~K\"{o}nighofer, S.~Niekum, and U.~Topcu, ``Safe reinforcement learning via shielding,'' in \emph{Proceedings of the Thirty-Second AAAI Conference on Artificial Intelligence and Thirtieth Innovative Applications of Artificial Intelligence Conference and Eighth AAAI Symposium on Educational Advances in Artificial Intelligence}, ser. AAAI'18/IAAI'18/EAAI'18.\hskip 1em plus 0.5em minus 0.4em\relax AAAI Press, 2018.

\bibitem{9157966}
P.~Jagtap, S.~Soudjani, and M.~Zamani, ``Formal synthesis of stochastic systems via control barrier certificates,'' \emph{IEEE Transactions on Automatic Control}, vol.~66, no.~7, pp. 3097--3110, 2021.

\bibitem{DBLP:journals/arcras/HsuHF24}
K.~Hsu, H.~Hu, and J.~F. Fisac, ``The safety filter: {A} unified view of safety-critical control in autonomous systems,'' \emph{Annu. Rev. Control. Robotics Auton. Syst.}, vol.~7, no.~1, 2024.

\bibitem{10745554}
A.~Zolfagharian, M.~Abdellatif, L.~C. Briand, and R.~S, ``{ SMARLA: A Safety Monitoring Approach for Deep Reinforcement Learning Agents },'' \emph{IEEE Transactions on Software Engineering}, vol.~51, no.~01, pp. 82--105, Jan. 2025.

\bibitem{DBLP:conf/rv/MallickGBD23}
S.~Mallick, S.~Ghosal, A.~Balakrishnan, and J.~Deshmukh, ``Safety monitoring for pedestrian detection in adverse conditions,'' in \emph{Runtime Verification - 23rd International Conference, {RV} 2023, Thessaloniki, Greece, October 3-6, 2023, Proceedings}, ser. Lecture Notes in Computer Science, P.~Katsaros and L.~Nenzi, Eds., vol. 14245.\hskip 1em plus 0.5em minus 0.4em\relax Springer, 2023, pp. 389--399.

\bibitem{10113472}
I.~Mitsioni, P.~Tajvar, D.~Kragic, J.~Tumova, and C.~Pek, ``Safe data-driven model predictive control of systems with complex dynamics,'' \emph{IEEE Transactions on Robotics}, vol.~39, no.~4, pp. 3242--3258, 2023.

\bibitem{10886383}
L.~Zhang, G.~Pantazis, S.~Han, and S.~Grammatico, ``An efficient risk-aware branch mpc for automated driving that is robust to uncertain vehicle behaviors,'' in \emph{2024 IEEE 63rd Conference on Decision and Control (CDC)}, 2024, pp. 8207--8212.

\bibitem{RAMAKRISHNA2020101760}
S.~Ramakrishna, C.~Harstell, M.~P. Burruss, G.~Karsai, and A.~Dubey, ``Dynamic-weighted simplex strategy for learning enabled cyber physical systems,'' \emph{Journal of Systems Architecture}, vol. 111, p. 101760, 2020.

\bibitem{10886150}
X.~Li, H.~E. Tseng, A.~Girard, and I.~Kolmanovsky, ``Autonomous driving with perception uncertainties: Deep-ensemble based adaptive cruise control,'' in \emph{2024 IEEE 63rd Conference on Decision and Control (CDC)}, 2024, pp. 8186--8192.

\bibitem{Tong2025EnsembleNN}
J.~Tong, S.~Du, and W.~Fan, ``Ensemble neural network‐based approximate model predictive control with strict guarantees,'' \emph{International Journal of Robust and Nonlinear Control}, vol.~35, pp. 7295 -- 7308, 2025.

\bibitem{ulgen}
B.~Yalcinkaya, H.~Torfah, A.~Desai, and S.~A. Seshia, ``Ulgen: A runtime assurance framework for programming safe cyber–physical systems,'' \emph{IEEE Transactions on Computer-Aided Design of Integrated Circuits and Systems}, vol.~42, no.~11, pp. 3679--3692, 2023.

\bibitem{FilippiCGS10}
S.~Filippi, O.~Capp{\'{e}}, A.~Garivier, and C.~Szepesv{\'{a}}ri, ``Parametric bandits: The generalized linear case,'' in \emph{Advances in Neural Information Processing Systems 23: 24th Annual Conference on Neural Information Processing Systems 2010. Proceedings of a meeting held 6-9 December 2010, Vancouver, British Columbia, Canada}, J.~D. Lafferty, C.~K.~I. Williams, J.~Shawe{-}Taylor, R.~S. Zemel, and A.~Culotta, Eds.\hskip 1em plus 0.5em minus 0.4em\relax Curran Associates, Inc., 2010, pp. 586--594.

\bibitem{FauryAJC22}
L.~Faury, M.~Abeille, K.~Jun, and C.~Calauz{\`{e}}nes, ``Jointly efficient and optimal algorithms for logistic bandits,'' in \emph{International Conference on Artificial Intelligence and Statistics, {AISTATS} 2022, 28-30 March 2022, Virtual Event}, ser. Proceedings of Machine Learning Research, G.~Camps{-}Valls, F.~J.~R. Ruiz, and I.~Valera, Eds., vol. 151.\hskip 1em plus 0.5em minus 0.4em\relax {PMLR}, 2022, pp. 546--580.

\bibitem{CarlssonBJD24}
E.~Carlsson, D.~Basu, F.~D. Johansson, and D.~P. Dubhashi, ``Pure exploration in bandits with linear constraints,'' in \emph{International Conference on Artificial Intelligence and Statistics, 2-4 May 2024, Palau de Congressos, Valencia, Spain}, ser. Proceedings of Machine Learning Research, S.~Dasgupta, S.~Mandt, and Y.~Li, Eds., vol. 238.\hskip 1em plus 0.5em minus 0.4em\relax {PMLR}, 2024, pp. 334--342.

\bibitem{saesae}
{SAE J3016}, ``On-road automated vehicle standards committee. {T}axonomy and definitions for terms related to on-road motor vehicle automated driving systems,'' p.~1, 2014.

\bibitem{checkingfinitetraces}
B.~Finkbeiner and H.~Sipma, ``Checking finite traces using alternating automata,'' \emph{Form. Methods Syst. Des.}, vol.~24, no.~2, pp. 101--127, Mar. 2004.

\bibitem{DBLP:conf/formats/MalerN04}
O.~Maler and D.~Nickovic, ``Monitoring temporal properties of continuous signals,'' in \emph{Formal Techniques, Modelling and Analysis of Timed and Fault-Tolerant Systems, Joint International Conferences on Formal Modelling and Analysis of Timed Systems, {FORMATS} 2004 and Formal Techniques in Real-Time and Fault-Tolerant Systems, {FTRTFT} 2004, Grenoble, France, September 22-24, 2004, Proceedings}, ser. Lecture Notes in Computer Science, Y.~Lakhnech and S.~Yovine, Eds., vol. 3253.\hskip 1em plus 0.5em minus 0.4em\relax Springer, 2004, pp. 152--166.

\bibitem{logisticreg}
\emph{Introduction to the Logistic Regression Model}.\hskip 1em plus 0.5em minus 0.4em\relax John Wiley \& Sons, Ltd, 2013, ch.~1, pp. 1--33.

\bibitem{faury2020improved}
L.~Faury, M.~Abeille, C.~Calauz{\`e}nes, and O.~Fercoq, ``Improved optimistic algorithms for logistic bandits,'' in \emph{International Conference on Machine Learning}.\hskip 1em plus 0.5em minus 0.4em\relax PMLR, 2020, pp. 3052--3060.

\bibitem{kveton2020randomized}
B.~Kveton, M.~Zaheer, C.~Szepesvari, L.~Li, M.~Ghavamzadeh, and C.~Boutilier, ``Randomized exploration in generalized linear bandits,'' in \emph{International Conference on Artificial Intelligence and Statistics}.\hskip 1em plus 0.5em minus 0.4em\relax PMLR, 2020, pp. 2066--2076.

\bibitem{shermanmorrison}
J.~Sherman and W.~J. Morrison, ``Adjustment of an inverse matrix corresponding to a change in one element of a given matrix,'' \emph{The Annals of Mathematical Statistics}, vol.~21, no.~1, pp. 124--127, 1950.

\bibitem{scenic}
D.~J. Fremont, T.~Dreossi, S.~Ghosh, X.~Yue, A.~L. Sangiovanni-Vincentelli, and S.~A. Seshia, ``Scenic: a language for scenario specification and scene generation,'' in \emph{Proceedings of the 40th ACM SIGPLAN Conference on Programming Language Design and Implementation}, ser. PLDI 2019.\hskip 1em plus 0.5em minus 0.4em\relax New York, NY, USA: Association for Computing Machinery, 2019, p. 63–78.

\bibitem{10.1007/978-3-030-53288-8_6}
D.~J. Fremont, J.~Chiu, D.~D. Margineantu, D.~Osipychev, and S.~A. Seshia, ``Formal analysis and redesign of a neural network-based aircraft taxiing system with verifai,'' in \emph{Computer Aided Verification: 32nd International Conference, CAV 2020, Los Angeles, CA, USA, July 21–24, 2020, Proceedings, Part I}.\hskip 1em plus 0.5em minus 0.4em\relax Berlin, Heidelberg: Springer-Verlag, 2020, p. 122–134.

\bibitem{DBLP:conf/aaai/Chen0Z0W24}
C.~Chen, J.~Liu, C.~Zhou, J.~Tang, and G.~Wu, ``Sketch and refine: Towards fast and accurate lane detection,'' in \emph{Thirty-Eighth {AAAI} Conference on Artificial Intelligence, {AAAI} 2024, Thirty-Sixth Conference on Innovative Applications of Artificial Intelligence, {IAAI} 2024, Fourteenth Symposium on Educational Advances in Artificial Intelligence, {EAAI} 2014, February 20-27, 2024, Vancouver, Canada}, M.~J. Wooldridge, J.~G. Dy, and S.~Natarajan, Eds.\hskip 1em plus 0.5em minus 0.4em\relax {AAAI} Press, 2024, pp. 1001--1009.

\bibitem{7995975}
Z.~Chen and X.~Huang, ``End-to-end learning for lane keeping of self-driving cars,'' in \emph{2017 IEEE Intelligent Vehicles Symposium (IV)}, 2017, pp. 1856--1860.

\bibitem{carla}
A.~Dosovitskiy, G.~Ros, F.~Codevilla, A.~Lopez, and V.~Koltun, ``{CARLA}: {An} open urban driving simulator,'' in \emph{Proceedings of the 1st Annual Conference on Robot Learning}, ser. Proceedings of Machine Learning Research, S.~Levine, V.~Vanhoucke, and K.~Goldberg, Eds., vol.~78.\hskip 1em plus 0.5em minus 0.4em\relax PMLR, 13--15 Nov 2017, pp. 1--16.

\bibitem{kirsch2022unifying}
A.~Kirsch and Y.~Gal, ``Unifying approaches in active learning and active sampling via fisher information and information-theoretic quantities,'' \emph{Transactions on Machine Learning Research}, 2022, expert Certification.

\end{thebibliography}

\clearpage
\newpage
\newpage

\appendix 

\section*{Appendix Outline}

In this appendix, we extend some results of our paper.

\begin{itemize}
    \item We provide details on how the epistemic uncertainty is computed in line 6 of \Cref{alg:active_learning}.
    \item We provide a complete proof for  \Cref{theorem:theorem1}.
    \item We provide additional details of the experiments performed.
    \item We include a plot illustrating the behavior of the monitoring policy in one simulation run.
    \item We include the \textsc{Scenic} program used to generate scenes for simulations.
\end{itemize}

\section{Epistemic Uncertainty} \label{section:uncertainty}
The key component of our approach is how to decide which controller to test during the learning phase. At every time $t$, the learner keeps track of the current MLE estimate of each controller, $\theta_{t,c}$,
\begin{align*}  
   \theta_{t,c} \in \argmax_\theta \sum_{j=1}^t \mathbb{1}_{\left(C_j = c \right)}  &( Y_j \log \sigma(\theta_c^\top \xi_j) + \nonumber \\ 
   & (1-Y_j)\log(1- \sigma(\theta_c^\top \xi_j)) ) \nonumber
\end{align*}

and the corresponding Hessian of the negative log-likelihood 
\begin{align}
    \Hess_{t}(\theta_{c, t}) = \sum_{s=1}^t \mathbf{I}_{(C_s = c)} \ \dot{\sigma}(\theta_{c, t}^\top \xi_s) \xi_s \xi_s^\top 
\label{eq:hess2}
\end{align}
where $\xi_s$ is the context observed in round $s$, and $\dot{\sigma}(.)=\sigma(.)\left(1 - \sigma(.)\right)$ is the derivative of $\sigma(.)$. 

At each round, we choose to test a controller and a context that satisfy 
\begin{align}\label{eq:epistemic-uncertainty}
    C_t \in \argmax\limits_{c \in \mathcal{C}} \|\xi_{t+1}\|_{\Hess^{-1}_{t}(\theta_{c, t})} & & \xi_t \in \argmax_{\xi \in \mathbb D^{\vcont}} \|\xi\|_{\Hess^{-1}_{t}(\theta_{C_t, t})}
\end{align}
where $||x||_V = |\sqrt{x^\top V x}|$. The quantity $\|\xi_{t+1}\|_{\Hess^{-1}_{t}(\theta_{c, t})}$ measures our epistemic uncertainty about the performance of controller $c$ in the direction of the vector $\xi_{t+1}$ (see \cite{kirsch2022unifying,Das2024ActivePO}) and sampling criterion in \Cref{eq:epistemic-uncertainty} will output a controller and a context with the highest epistemic uncertainty.

The Hessian is updated using the Sherman-Morrison formula \cite{shermanmorrison}. The intuition behind using this formula is that it efficiently computes the inverse of a matrix that has been modified by just a rank-one update. This formula provides :
\begin{align*}
    H&_t^{-1}(\theta_{t-1}) = \\ & H_{t-1}^{-1}(\theta_{t-1}) - \dot\sigma (\xi_t^\top \theta_{t-1}) \frac{H_{t-1}^{-1}(\theta_{t-1})\xi_t \xi_t^\top H_{t-1}^{-1}(\theta_{t-1})}{1 + \dot\sigma(\xi_t^\top \theta_{t-1}) ||\xi_t||^2_{H_{t-1}^{-1}(\theta_{t-1})}}
\end{align*} 

\section{Proof of Theorem 1} 
\label{appendix:theorem1}

Notice that our goal is to learn an optimal policy $\pi$ such that it minimizes the regret:
\begin{align*}
\max \limits_{\xi \in \mathbb D^{\vcont}} |\sigma(\theta_{\pi(\xi)}^\top \xi) - \sigma(\theta_{\pi^*(\xi)}^\top \xi) | 
\end{align*}

This is, we want to find the vectors $\{\theta_c\}_{c \in C}$ that solve the minimization problem above. 

To derive theoretical bounds on the regret of our problem, we first derive bounds for a more complex problem and then use them to bound the regret.
We start by showing how to use padding to transform the previous problem into the problem of minimizing the regret for every controller, and not just for the safest one. This is, not only for the one that minimizes $\sigma(\theta_{c}^\top \xi)$.

After this transformation, we can apply \Cref{lemma:lemma1,lemma:elliptic} of the following sections to prove \Cref{theorem:theorem1}. 

\subsection{Conversion using padding}

Let $\xi \in \mathbb{D}^{V_{cont}}$ ($\subset \mathbb{R}^d$) a context  and $n=|\mathcal{C}|$ the number of controllers. For each controller $c_i \in \mathcal{C}$ with $1 \leq i \leq n$ we define $\xi^i \in \mathbb{R}^{n \cdot d}$ as follows: 
$$(\xi^i)_j = \left\{ \begin{array}{lcc} (\xi)_{j - (i-1)d} & \text{if} & (i-1)d+1 \leq j \leq id  \\  0 & \text{otherwise} \\ \end{array} \right.$$

Notice that, to generate vectors $\xi^i$, we strategically add 0-padding to the context vector $\xi$, and that computing the maximum likelihood estimate is equivalent to computing: 
\begin{align*}
   \theta_{T,c} \in \argmax_\theta \sum_{t=1}^T & \mathbb{1}_{(C_t = c_i)} ( Y_t \log \sigma(\theta^\top \xi_t^i) + \\ & (1-Y_t)(1-\sigma(\theta^\top \xi_t^i)) )
\end{align*}

At each step of \Cref{alg:active_learning} we can define the data point $\{\xi_t^{i_t},Y_t\}$ where $c_{i_t} = C_t$. We can define $\theta_{T} \in \mathbb{R}^{n\cdot d}$ such that its values between positions $(i-1)d+1$ and $id$ correspond to $\theta_{T,c_i}$. Because of the definition of $\xi^i$, computing $\theta_{T,c}$ for every controller $c$ is equivalent to compute:
\begin{align}
   \theta_{T} \in \argmax_\theta \sum_{t=1}^T & ( Y_t \log \sigma(\theta^\top \xi_t^{i_t}) + \nonumber \\ & (1-Y_t)(1-\sigma(\theta^\top \xi_t^{i_t})) )
   \label{eq:mle_estimator_appendix}
\end{align}

Assuming that $\theta_{T^*}$ converges to the optimal policy $\pi$ for a sufficiently large value $T^*$, the regret of our problem is then equivalent to:

\begin{align*}   
R := \max_{\xi \in \mathbb D^{\vcont}}  |\sigma(\theta_{*}^\top \xi^m) - \sigma(\theta_{T^*}^\top \xi^m)|  \\ 
\text{ where } c_m \in \argmin \limits_{c\in C} \sigma(\theta_{c}^\top \xi)
\end{align*}

We define now the following regret: 
\begin{align*}   
R(T)  := & \max_{\substack{\xi \in \mathbb D^{\vcont} \\ 1 \leq i \leq n}}  |\sigma(\theta_{*}^\top \xi^i) - \sigma(\theta_{T}^\top \xi^i)| <  \\ 
&\max_{\substack{\xi \in \mathbb D^{\vcont} \\ 1 \leq i \leq n}}  | \theta_{*}^\top \xi^i  - \theta_{T}^\top \xi^i|
\label{eq:regret}
\end{align*}

Where the inequality comes from $\sigma(\cdot)$ being Lipschitz with constant 1/4.

It is clear that $R \leq R(T^*)$. It remains to show that $R(T)$ conforms with the bounds as stated in \Cref{theorem:theorem1}.

\subsection{Results needed to prove \Cref{theorem:theorem1}} \label{section:auxiliary}

For each iteration of the bandits' algorithm, there is a bound on the error between the optimal monitor and the learned monitor. This is a consequence of Lemma 1 of \cite{Das2024ActivePO}.
\begin{lemma}[Estimation error at round $t$ \cite{Das2024ActivePO}]
    Let $\delta \in (0,1]$. Assuming that the optimal monitor is bounded, i.e., $||\theta_*|| \leq q$ for some $q>0$, let $\lambda = \frac{1}{4q^2(2+q)^2}$. Then, using \Cref{alg:active_learning} to train a monitor $\theta_t \in \mathbb{R}^{n \cdot d}$ at each round $t$, 
    with probability at least $1-\delta$, we have for some constant $b > 0$,
    \begin{align}
        ||\theta_* - \theta_t||_{H_t(\theta_t)} \leq bq^{1/2}\gamma_t(\delta) 
    \end{align}
    with $\gamma_t(\delta) = bq \sqrt{n d \log \frac{qt}{nd} + \log \frac{t}{\delta}}$, $||\theta_*||$ the Euclidean norm of $\theta_*$, 
    and $H_t(\theta_t)$ the Hessian as defined in \Cref{eq:hess2}.
    \label{lemma:lemma1}
\end{lemma}

The details of the proof can be found in \cite{Das2024ActivePO}.

The next Lemma allows us to derive the final bound of \Cref{theorem:theorem1}, as can be seen in the next section.

\begin{lemma}[Elliptic potential lemma \cite{lattimore2020bandit}]
    Let $\{x_s\}_{s=1}^t$ be a sequence of vectors in $\mathbb{R}^d$ such that $||x_s|| \leq L < \infty $ for any $s \in [t]$. Let $V_t = \sum_{s=1}^{t-1}x_s x_s^\top + \lambda I$. Then, 
    \begin{align*}
        \sum\limits_{s=1}^t ||x_s||^2_{V_s^{-1}} \leq 2d\log \big(1 + \frac{tL^2}{\lambda d}\big) 
    \end{align*}
    \label{lemma:elliptic}
\end{lemma}

\subsection{Proof of \Cref{theorem:theorem1}} \label{section:appendixproof}

We start with a reminder of Theorem 1:

    \textit{Using \Cref{alg:active_learning}, selecting the controller and context with the highest uncertainty,  we have that the regret decays with the number of rounds $T$ with order $\mathcal{O}(\sqrt{log(T)^2/T})$. }

In order to prove \Cref{theorem:theorem1}, we show that the following suboptimality gap $R(T,x) := |\theta_{*}^\top x - \theta_T^\top  x|$ is bounded for every (padded) context $x$. This is important because, by definition, $R(T) \leq \max\limits_x R(T,x)$. 

We show that $R(T,x)$ is bounded in terms of $||\theta_* - \theta_t||_{\Hess^{-1}_{t}(\theta_{t})}$
and in terms of $||x||_{\Hess^{-1}_{t}(\theta_{t})}$, the uncertainty of the context. This allows us to apply \Cref{lemma:lemma1,lemma:elliptic} to find bounds of $R(T,x)$ in terms of $T$.

Before proving \Cref{theorem:theorem1}, it is necessary to prove the following corollary, which follows from \Cref{lemma:lemma1}.

\begin{corollary}
Let $\delta \in (0,1]$. With probability at least $1-\delta$, \Cref{alg:active_learning} holds the following bound:
    $$R(T) \leq b \gamma_T(\delta) \sqrt{q \log(1+\frac{T}{\lambda \kappa n d}) \frac{\kappa n d}{T}}$$
for a constant $b>0$, $\gamma_t(\delta)$, $q$ and $\lambda$ as defined in \Cref{lemma:lemma1}, and $\kappa = \max\limits_{x \in \mathcal{X}} \max\limits_{\theta \in \Theta} \frac{1}{\dot\sigma(\theta^\top x)}$.
\label{corollary:corollary1}
\end{corollary}

In the experiments, we use the MLE (\Cref{eq:mle_estimator_appendix}) to estimate the policy $\pi$. For the proof of \Cref{corollary:corollary1}, we use $\theta_T = \frac{1}{T} \sum_{t=1}^T \theta_t$ to get the bounds on $R(T,x)$. This average of the MLE estimators quickly converges to the MLE for sufficiently large values of $T$, and will help us derive the theoretical bounds.

 For a (padded) context $x \in \mathbb{R}^{n\cdot d}$, we have that:
\begin{align*}
    R(T,x) & =  |\theta_{*}^\top x - \theta_T^\top  x|
    =  |(\theta_{*}^\top - \frac{1}{T} \sum\limits_{t=1}^T \theta_{t}^\top) x| \\ 
    & \leq  \frac{1}{T} \sum\limits_{t=1}^T |(\theta_{*}^\top - \theta_{t}^\top) x| \\ & (\text{Cauchy-Schwarz})  \\
    & \leq \frac{1}{T} \sum\limits_{t=1}^T ||\theta_{*}^\top -\theta_{t}^\top||_{H_t(\theta_t)} ||x||_{H_t^{-1}(\theta_t)} \\ & (\text{\Cref{lemma:lemma1}}) \\
    & \leq \frac{bq^{1/2}\gamma_T(\delta)}{T} \sum\limits_{t=1}^T  ||x||_{H_t^{-1} (\theta_t)} \\  
    & \leq \frac{bq^{1/2}\gamma_T(\delta)}{T} \sum\limits_{t=1}^T  ||x_{t}||_{H_t^{-1} (\theta_t)}
\end{align*}

We use that $x_t \in \argmax_{x \in \mathcal{X}} ||x||_{\Hess^{-1}_{t}(\theta_{t})}$ to get the last inequality, this is, $x_t$ is the context with the highest uncertainty.

Now, because $ V_t \preceq \kappa H_t(\theta_t)$, we have that we can upper bound $||x_{t}||_{H_t^{-1} (\theta_t)}$ by $\sqrt{\kappa} ||x_{t}||_{V_t^{-1}}$. For this reason, we can apply \Cref{lemma:elliptic}. Then,

\begin{align*}
    R(T,x) &  \leq \frac{b\sqrt{\kappa q}\gamma_T(\delta)}{T} \sum\limits_{t=1}^T  ||x_{t}||_{V_t^{-1}} \\ & (\text{Cauchy-Schwarz})  \\
    & \leq \frac{b\sqrt{\kappa q}\gamma_T(\delta)}{T} \sqrt{T \sum\limits_{t=1}^T  ||x_{t}||^2_{V_t^{-1}}} \\ & (\text{\Cref{lemma:elliptic}})  \\
    & \leq \frac{b\sqrt{\kappa q}\gamma_T(\delta)}{T} \sqrt{2ndT \log \Big(1 + \frac{T}{\lambda \kappa n d} \Big)}  \\ & (\text{Def. of $\gamma_T(\delta)$})  \\
    & = b q^{3/2} \sqrt{\Big(nd \log (\frac{qT}{nd}) + \log(\frac{T}{\delta}) \Big)} \ \cdot \\ & \ \ \ \ \ \ \ \ \ \ \ \ \ \ \ \ \ \ \ \  \sqrt{ \log \Big( 1 + \frac{T}{\lambda \kappa n d} \Big) \frac{\kappa n d}{T} \Big)}
\end{align*}

Finally,because $R(T) \leq \max\limits_{x} R(T,x)$, and seeing that in the last inequality $b$, $q$, $n$, $d$, $\delta$, $\lambda$, and $\kappa$ are constants, we can deduce \Cref{theorem:theorem1}.

\section{Experiments} \label{appendix:experiments}

In this section, we explain the missing details related to the experiments, and we include an example of a simulation, where we show how the monitor works in practice.

\subsection{CNN Structure}

The CNN controllers have the structure shown in \Cref{CNNStructure}: three convolutional layers, with two fully connected layers at the end. Depending on the scenario, the output will be given by $N=1$ (CTE) or $N=2$ neurons (CTE, target speed).

\subsection{Controllers of Experiment Contextual vs Bagging}

The 15 biased controllers are trained each on a specific context $\xi = (w,t,i,d)$ such that $(w,t)$ is given by one of the weather presets $W = \{$\emph{ClearNoon, ClearSunset, HardRainNoon, HardRainSunset, CloudyNoon}$\}$, and $d$ take values of $d=100$ (no other car ahead), $d \in [5,10]$, or $d \in [20, 30]$. 

\subsection{Contextual vs. Boosting}
\label{section:boosting}
\begin{figure}
\centering

   \scalebox{0.8}{
    \begin{tikzpicture}
    \begin{axis}[
            symbolic x coords={{Boosting}, {Boosting weaker learners},
            },
            enlarge x limits={abs=1.5cm},
            width=9cm,
            height=6.7cm,
            xtick=data,
            ylabel={Average reward},
            ymin=0.0, ymax=1,
            ytick={0.1,0.2,0.3,0.4,0.5,0.6,0.7,0.8,0.9,1.0},
            ybar,
            bar width=7pt,
            legend image code/.code={
                \draw [#1] (0cm,-0.1cm) rectangle (0.2cm,0.25cm); },
            legend style={at={(0.5,-0.2)},anchor=north,legend columns=-1},
        ]
        \addplot coordinates {(Boosting, 0.076)(Boosting weaker learners, 0.097)
        }; 
        \addplot+[error bars/.cd, y dir=both,y explicit] coordinates {(Boosting, 0.148)(Boosting weaker learners, 0.111)
        }; 
        \addplot+[error bars/.cd, y dir=both,y explicit] coordinates {(Boosting, 0.176)(Boosting weaker learners, 0.133)
        }; 
        \addplot+[error bars/.cd, y dir=both,y explicit] coordinates {(Boosting, 0.220)(Boosting weaker learners, 0.144)
        }; 
        \addplot+[error bars/.cd, y dir=both,y explicit] coordinates {(Boosting, 0.304)(Boosting weaker learners, 0.198)
        }; 
        \addplot+[error bars/.cd, y dir=both,y explicit] coordinates {(Boosting, 0.488)(Boosting weaker learners, 0.287)
        }; 

        \legend{Non-Contextual,FP $0\%$, FP $5\%$,FP $10\%$,FP $20\%$,FP $30\%$,Weak learners}

        \end{axis}
    \end{tikzpicture}
}
    \caption{Extension of RQ2: Contextual vs. Boosting 
    }
    \label{fig:extensionboosting}
\end{figure}
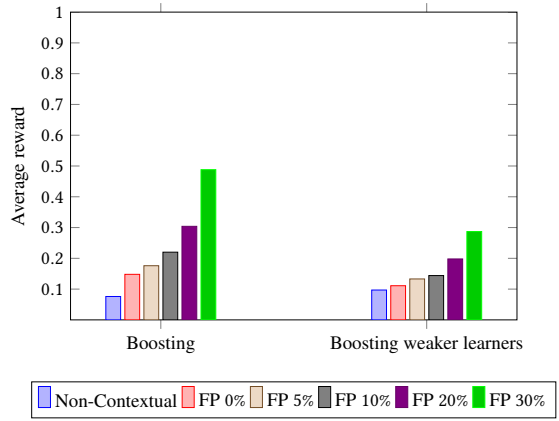

We further used adaptive boosting to train an ensemble of 15 controllers for Scenario 1. This ensemble did not perform well, achieving much worse performance than any other model. The main reason is that the first controller trained learns with high accuracy, and the rest overfit on the few data points for which the first is poor. As a consequence, 14 out of 15 controllers do not generalize well, and the performance of the first controller is diluted due to the influence of the others  (see \Cref{fig:extensionboosting}).
We evaluated a second boosting ensemble using weaker learners on 100 simulations and obtained similar results. For these reasons, we decided not to use them as baselines.

We designed another boosting ensemble using weaker learners, specifically using smaller CNNs. The goal was to determine whether the controllers would overfit even when using weaker learners that would perform worse individually. The results indicate that the models still overfit, and because the controllers are now less complex, the performance of the monitor is worse. 

\subsection{Example behavior of monitor along a simulation} \label{appendix:simulation}

\begin{figure}[t]
    \centering
\scalebox{0.8}{
\begin{tikzpicture}
    \centering
    \begin{axis}[height=5cm,width=10cm, tick align=outside, tickpos=left,
            ylabel={Distance},
            xticklabel=\empty,
            xmin=-5, xmax=300,
            ymin=10, ymax=105,
            legend style={at={(0.05,0.6)},anchor=north west},
            ymajorgrids=false,
        ]
        \addplot[color=blue] coordinates {(0,19.29077368458797)(1,19.37511057389816)(2,19.468985697729043)(3,19.57289393373366)(4,19.681833565789905)(5,19.79431305802013)(6,19.911562570521017)(7,20.034541310129967)(8,20.162558589216687)(9,20.29562861587874)(10,20.433280934082205)(11,20.5755507683742)(12,20.723152038466143)(13,20.87649113601092)(14,21.03581456072658)(15,21.20090492317929)(16,21.371980172557688)(17,21.549166860687187)(18,21.732828900835813)(19,21.92241985065801)(20,22.116746930339616)(21,22.317014341985107)(22,22.523722132940307)(23,22.736275015923795)(24,22.954933385412943)(25,23.179859158053667)(26,23.41156133396133)(27,23.64988847648909)(28,23.89428125571725)(29,24.14520934698162)(30,24.40213309596595)(31,24.66047465220586)(32,24.918252607779905)(33,25.173478804218156)(34,25.424421680335005)(35,25.670007154063573)(36,25.909001531767675)(37,26.14068548287789)(38,26.365506059340422)(39,26.583365743059435)(40,26.79644139694445)(41,27.00513014812044)(42,27.20947336301086)(43,27.41062594443366)(44,27.60893715003189)(45,27.804926675305783)(46,27.99924987786204)(47,28.19206434864878)(48,28.385129949339493)(49,28.57893434067725)(50,28.772478025594435)(51,28.96664598591213)(52,29.1593848334588)(53,29.350721698658862)(54,29.541047369352064)(55,29.73109801853432)(56,29.92298192278509)(57,30.114804845547248)(58,30.30769205184548)(59,30.50166023480902)(60,30.697188118788763)(61,30.894486610836424)(62,31.092036554724626)(63,31.29156614151425)(64,31.490969715338974)(65,31.691538583628844)(66,31.89232398019968)(67,32.092005468316245)(68,32.29196556510768)(69,32.49095580096267)(70,32.6896102862789)(71,32.88652633100278)(72,33.08338721569513)(73,33.27877317236927)(74,33.47375913455144)(75,33.66913344824107)(76,33.86477000309336)(77,34.060474097295824)(78,34.25627940464682)(79,34.453319948947275)(80,34.65037122204139)(81,34.84804351974003)(82,35.0458059281655)(83,35.24447040628254)(84,35.4411014291863)(85,35.63795457250408)(86,35.83479939719492)(87,36.03252135085943)(88,36.23037593178636)(89,36.42846457590518)(90,36.626102216531336)(91,36.82446195376046)(92,37.02160263298468)(93,37.21811311011983)(94,37.41452066729046)(95,37.611961843126494)(96,37.80922977812111)(97,38.00634266745408)(98,38.203266696104784)(99,38.40056668802295)(100,38.598736931710825)(101,38.79662402680106)(102,38.99428431732552)(103,39.19025165847362)(104,39.38616807510203)(105,39.58240269674482)(106,39.77897351295904)(107,39.97542807574035)(108,40.17297964200515)(109,40.37067805710316)(110,40.56840637558805)(111,40.765996254711766)(112,40.963831033077135)(113,41.162590425866256)(114,41.36061669514889)(115,41.558783222190456)(116,41.75740130362277)(117,41.95544315474926)(118,42.15341602809814)(119,42.35148529897247)(120,42.54916324170036)(121,42.74686101718401)(122,42.94445935431258)(123,43.141980697314565)(124,43.33968541582531)(125,43.53709288833631)(126,43.73469727974042)(127,43.93180204935882)(128,44.12923263790852)(129,44.326393395400245)(130,44.52363177370635)(131,44.721106568085126)(132,44.91930266668514)(133,45.11761594678588)(134,45.3156550043171)(135,45.51281789477096)(136,45.71004951161192)(137,45.908345235467166)(138,46.10462450766765)(139,46.30010991348077)(140,46.496623565616126)(141,46.69345879140261)(142,46.890991457544246)(143,47.08834684213246)(144,47.285655585640434)(145,47.48494235242953)(146,47.68189142687962)(147,47.87902594033875)(148,48.07575997022221)(149,48.272613877698014)(150,48.47039589106189)(151,48.667427056530734)(152,48.86488549201805)(153,49.06270301265904)(154,49.26031196785045)(155,49.45799562160452)(156,49.655033908409)(157,49.85133633701488)(158,100.0)(159,100.0)(160,100.0)(161,100.0)(162,100.0)(163,100.0)(164,100.0)(165,100.0)(166,100.0)(167,100.0)(168,100.0)(169,100.0)(170,100.0)(171,100.0)(172,100.0)(173,100.0)(174,100.0)(175,100.0)(176,100.0)(177,100.0)(178,100.0)(179,100.0)(180,100.0)(181,100.0)(182,100.0)(183,100.0)(184,100.0)(185,100.0)(186,100.0)(187,100.0)(188,100.0)(189,100.0)(190,100.0)(191,100.0)(192,100.0)(193,100.0)(194,100.0)(195,100.0)(196,100.0)(197,100.0)(198,100.0)(199,100.0)(200,100.0)(201,100.0)(202,100.0)(203,100.0)(204,100.0)(205,100.0)(206,100.0)(207,100.0)(208,100.0)(209,100.0)(210,100.0)(211,100.0)(212,100.0)(213,100.0)(214,100.0)(215,100.0)(216,100.0)(217,100.0)(218,100.0)(219,100.0)(220,100.0)(221,100.0)(222,100.0)(223,100.0)(224,100.0)(225,100.0)(226,100.0)(227,100.0)(228,100.0)(229,100.0)(230,100.0)(231,100.0)(232,100.0)(233,100.0)(234,100.0)(235,100.0)(236,100.0)(237,100.0)(238,100.0)(239,100.0)(240,100.0)(241,100.0)(242,100.0)(243,100.0)(244,100.0)(245,100.0)(246,100.0)(247,100.0)(248,100.0)(249,100.0)(250,100.0)(251,100.0)(252,100.0)(253,100.0)(254,100.0)(255,100.0)(256,100.0)(257,100.0)(258,100.0)(259,100.0)(260,100.0)(261,100.0)(262,100.0)(263,100.0)(264,100.0)(265,100.0)(266,100.0)(267,100.0)(268,100.0)(269,100.0)(270,100.0)(271,100.0)(272,100.0)(273,100.0)(274,100.0)(275,100.0)(276,100.0)(277,100.0)(278,100.0)(279,100.0)(280,100.0)(281,100.0)(282,100.0)(283,100.0)(284,100.0)(285,100.0)(286,100.0)(287,100.0)(288,100.0)(289,100.0)(290,100.0)(291,100.0)(292,100.0)(293,100.0)(294,100.0)(295,100.0)(296,100.0)(297,100.0)(298,100.0)(299,100.0)(300,100.0)};
        \end{axis}  
        \begin{axis}[yshift=-3.8cm,height=5cm,width=10cm,
            ylabel={Intersection},
            xticklabel=\empty,
            xmin=-5, xmax=300,
            ymin=-0.05, ymax=1.05,
            legend style={at={(0.05,0.6)},anchor=north west},
            ytick={0,1.0},
            ymajorgrids=false,
        ]

        \addplot[color=orange] coordinates {(0,0)(1,0)(2,0)(3,0)(4,0)(5,0)(6,0)(7,0)(8,0)(9,0)(10,0)(11,0)(12,0)(13,0)(14,0)(15,0)(16,0)(17,0)(18,0)(19,0)(20,0)(21,0)(22,0)(23,0)(24,0)(25,0)(26,0)(27,0)(28,0)(29,0)(30,0)(31,0)(32,0)(33,0)(34,0)(35,0)(36,0)(37,0)(38,0)(39,0)(40,0)(41,0)(42,0)(43,0)(44,0)(45,0)(46,0)(47,0)(48,0)(49,0)(50,0)(51,0)(52,0)(53,0)(54,0)(55,0)(56,0)(57,0)(58,0)(59,0)(60,0)(61,0)(62,0)(63,0)(64,0)(65,0)(66,0)(67,0)(68,1)(69,1)(70,1)(71,1)(72,1)(73,1)(74,1)(75,1)(76,1)(77,1)(78,1)(79,1)(80,1)(81,1)(82,1)(83,1)(84,1)(85,1)(86,1)(87,1)(88,1)(89,1)(90,1)(91,1)(92,1)(93,1)(94,1)(95,1)(96,1)(97,1)(98,1)(99,1)(100,1)(101,1)(102,1)(103,1)(104,1)(105,1)(106,1)(107,1)(108,1)(109,1)(110,1)(111,1)(112,1)(113,1)(114,1)(115,1)(116,1)(117,1)(118,1)(119,1)(120,1)(121,1)(122,1)(123,1)(124,1)(125,1)(126,1)(127,1)(128,1)(129,1)(130,1)(131,1)(132,1)(133,1)(134,1)(135,1)(136,1)(137,1)(138,1)(139,1)(140,1)(141,1)(142,1)(143,1)(144,1)(145,1)(146,1)(147,1)(148,1)(149,1)(150,1)(151,0)(152,0)(153,0)(154,0)(155,0)(156,0)(157,0)(158,0)(159,0)(160,0)(161,0)(162,0)(163,0)(164,0)(165,0)(166,0)(167,0)(168,0)(169,0)(170,0)(171,0)(172,0)(173,0)(174,0)(175,0)(176,0)(177,0)(178,0)(179,0)(180,0)(181,0)(182,0)(183,0)(184,0)(185,0)(186,0)(187,0)(188,0)(189,0)(190,0)(191,0)(192,0)(193,0)(194,0)(195,0)(196,0)(197,0)(198,0)(199,0)(200,0)(201,0)(202,0)(203,0)(204,0)(205,0)(206,0)(207,0)(208,0)(209,0)(210,0)(211,0)(212,0)(213,0)(214,0)(215,0)(216,0)(217,0)(218,0)(219,0)(220,0)(221,0)(222,0)(223,0)(224,0)(225,0)(226,0)(227,0)(228,0)(229,0)(230,0)(231,0)(232,0)(233,0)(234,0)(235,0)(236,0)(237,0)(238,0)(239,0)(240,0)(241,0)(242,0)(243,0)(244,0)(245,0)(246,0)(247,0)(248,0)(249,0)(250,0)(251,0)(252,0)(253,0)(254,0)(255,0)(256,0)(257,0)(258,0)(259,0)(260,0)(261,0)(262,0)(263,0)(264,0)(265,0)(266,0)(267,0)(268,0)(269,0)(270,0)(271,0)(272,0)(273,0)(274,0)(275,0)(276,0)(277,0)(278,0)(279,0)(280,0)(281,0)(282,0)(283,0)(284,0)(285,0)(286,0)(287,0)(288,0)(289,0)(290,0)(291,0)(292,0)(293,0)(294,0)(295,0)(296,0)(297,0)(298,0)(299,0)(300,0)};

    \end{axis}
    \begin{axis}[yshift=-7.6cm,height=5cm,width=10cm,
            ylabel={Controller},
            xlabel={Timesteps},
            xmin=-5, xmax=300,
            ymin=-0.05, ymax=15.05,
            legend style={at={(0.05,0.6)},anchor=north west},
            ytick={0,1,2,3,4,5,6,7,8,9,10,11,12,13,14},
            yticklabels={$c_0$,$c_1$,$c_2$,$c_3$,$c_4$,$c_5$,$c_6$,$c_7$,$c_8$,$c_9$,$c_{10}$,$c_{11}$,$c_{12}$,$c_{13}$,$c_{14}$},
            ymajorgrids=false,
        ]

        \addplot[color=black] coordinates {(0,9)(1,9)(2,9)(3,9)(4,9)(5,9)(6,9)(7,9)(8,9)(9,9)(10,9)(11,9)(12,9)(13,9)(14,9)(15,9)(16,9)(17,9)(18,9)(19,9)(20,9)(21,9)(22,9)(23,9)(24,9)(25,9)(26,9)(27,9)(28,9)(29,9)(30,9)(31,9)(32,9)(33,9)(34,9)(35,9)(36,9)(37,9)(38,9)(39,9)(40,9)(41,9)(42,9)(43,9)(44,9)(45,9)(46,9)(47,9)(48,9)(49,9)(50,9)(51,9)(52,9)(53,9)(54,9)(55,9)(56,9)(57,9)(58,9)(59,9)(60,9)(61,9)(62,9)(63,9)(64,9)(65,9)(66,9)(67,9)(68,5)(69,5)(70,5)(71,5)(72,5)(73,5)(74,5)(75,5)(76,5)(77,5)(78,5)(79,5)(80,5)(81,5)(82,5)(83,5)(84,5)(85,5)(86,5)(87,5)(88,5)(89,5)(90,5)(91,5)(92,5)(93,5)(94,5)(95,5)(96,5)(97,5)(98,5)(99,5)(100,5)(101,5)(102,5)(103,5)(104,5)(105,5)(106,5)(107,5)(108,5)(109,5)(110,5)(111,5)(112,5)(113,5)(114,5)(115,5)(116,5)(117,5)(118,5)(119,5)(120,5)(121,5)(122,5)(123,5)(124,5)(125,5)(126,5)(127,5)(128,5)(129,5)(130,5)(131,5)(132,5)(133,5)(134,5)(135,5)(136,5)(137,5)(138,5)(139,5)(140,5)(141,5)(142,5)(143,5)(144,5)(145,5)(146,5)(147,5)(148,5)(149,5)(150,5)(151,9)(152,9)(153,9)(154,9)(155,9)(156,9)(157,9)(158,10)(159,10)(160,10)(161,10)(162,10)(163,10)(164,10)(165,10)(166,10)(167,10)(168,10)(169,10)(170,10)(171,10)(172,10)(173,10)(174,10)(175,10)(176,10)(177,10)(178,10)(179,10)(180,10)(181,10)(182,10)(183,10)(184,10)(185,10)(186,10)(187,10)(188,10)(189,10)(190,10)(191,10)(192,10)(193,10)(194,10)(195,10)(196,10)(197,10)(198,10)(199,10)(200,10)(201,10)(202,10)(203,10)(204,10)(205,10)(206,10)(207,10)(208,10)(209,10)(210,10)(211,10)(212,10)(213,10)(214,10)(215,10)(216,10)(217,10)(218,10)(219,10)(220,10)(221,10)(222,10)(223,10)(224,10)(225,10)(226,10)(227,10)(228,10)(229,10)(230,10)(231,10)(232,10)(233,10)(234,10)(235,10)(236,10)(237,10)(238,10)(239,10)(240,10)(241,10)(242,10)(243,10)(244,10)(245,10)(246,10)(247,10)(248,10)(249,10)(250,10)(251,10)(252,10)(253,10)(254,10)(255,10)(256,10)(257,10)(258,10)(259,10)(260,10)(261,10)(262,10)(263,10)(264,10)(265,10)(266,10)(267,10)(268,10)(269,10)(270,10)(271,10)(272,10)(273,10)(274,10)(275,10)(276,10)(277,10)(278,10)(279,10)(280,10)(281,10)(282,10)(283,10)(284,10)(285,10)(286,10)(287,10)(288,10)(289,10)(290,10)(291,10)(292,10)(293,10)(294,10)(295,10)(296,10)(297,10)(298,10)(299,10)(300,10)};

    \end{axis}
\end{tikzpicture}
}
\caption{Example of a simulation, where a monitor of Use Case 1 and Setting $S_3$ (\Cref{fig:setting3}) switches between controllers based on the weather, the distance to a car ahead, and the road type.} \label{fig:switch}
\end{figure}
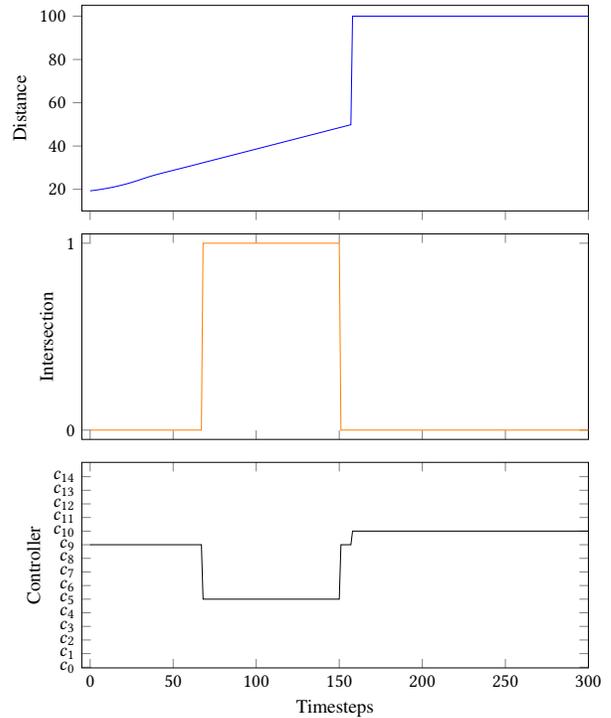

In \Cref{fig:switch}, we use one of the testing simulations to illustrate how our monitors work. The weather during this simulation is \emph{WetNoon}. We can observe that, at the beginning, the ego car is on a straight road (intersection=0) with another car around 20 meters ahead (distance=20). The ego car uses first the controller $c_9$. Once it arrives at an intersection, after around 60 time steps, it switches to the controller $c_5$. Finally, shortly after leaving the intersection, the ego car stops seeing the car ahead, represented by distance=$100$. The monitor switches momentarily to the controller $c_9$ before switching to the controller $c_{10}$ after leaving the intersection. 

During the simulation, the lowest confidence of the monitor on one of the controllers is 0.9783. If we had used a confidence threshold higher than that value, the monitor would have switched to the safe controller instead.

\section{Scenic Program} \label{appendix:scenic}

The \textsc{Scenic} program of \Cref{scenicProgram1} creates scenes for simulations where there is another car ahead of the ego car. The program generates a scene choosing one of the 14 weather presets available in \textsc{Carla}. Then, it places the ego car in the center of a random lane. Finally, it places another car in front of the ego car with a random distance between 5 and 50. The cars will then drive at constant speed. The monitor is incorporated in the behavior of the ego car: EgoBehavior.

\begin{figure*}
\begin{center}
\begin{minipage}{13cm}
\centering
\begin{verbatim}

class convNet(torch.nn.Module):
    def __init__(self):
        super(convNet, self).__init__()

        self.model = torch.nn.Sequential(
            torch.nn.Conv2d(3, 32, kernel_size=3, stride=2),
            torch.nn.LeakyReLU(),
            torch.nn.MaxPool2d(kernel_size=2),
            torch.nn.Conv2d(32, 64, kernel_size=3, stride=2),
            torch.nn.LeakyReLU(),
            torch.nn.MaxPool2d(kernel_size=2),
            torch.nn.Conv2d(64, 128, kernel_size=3, stride=2),
            torch.nn.Flatten(),
            
            self.fc1 = torch.nn.Linear(1025,1024)
            self.head = torch.nn.Linear(1024, N)
        )



\end{verbatim}
\end{minipage}
\end{center}
\caption{Definition of the CNNs used as controllers.}
\label{CNNStructure}
\end{figure*}

\begin{figure*}
\begin{center}
\begin{minipage}{13cm}
\centering
\begin{verbatim}

## SET MAP AND MODEL 
param map = localPath('maps/Town01.xodr')
param carla_map = 'Town01'


## CONSTANTS
EGO_MODEL = "vehicle.tesla.model3"
EGO_SPEED = 5
CAR1_SPEED = Range(3,7)
DIST_CAR1 = Range(5, 50) 


## STATIC FEATURES
param weather = Uniform('ClearNoon','CloudyNoon','WetNoon',
  'WetCloudyNoon','MidRainyNoon', 'HardRainNoon', 'SoftRainNoon',
  'ClearSunset', 'CloudySunset', 'WetSunset', 'WetCloudySunset', 
  'MidRainSunset', 'HardRainSunset', 'SoftRainSunset')



## DEFINING BEHAVIORS
behavior EgoBehavior(speed=10):
    action = LaneKSimplex("simulation_data")
    take action
    while True: 
        try:
            take action
        interrupt when self.distanceToClosest(Object) < 6:
            take SetBrakeAction(1), SetThrottleAction(0)

## DEFINING SPATIAL RELATIONS

lane = Uniform(*network.lanes)
start = new OrientedPoint on lane.centerline
ego = new SimplexCarlaCar at start,
    with blueprint EGO_MODEL,
    with behavior EgoBehavior(EGO_SPEED))

car1 = new Car following roadDirection from start for DIST_CAR1,
            with behavior FollowLaneBehavior(CAR1_SPEED)

require ego can see car1

terminate when ego.collision > 0 and ego.speed < 0.1 
                and (distance to start) > 1


\end{verbatim}
\end{minipage}
\end{center}
\caption{Example of \textsc{Scenic} program for Use Case 1: Lane Keeping.}
\label{scenicProgram1}
\end{figure*}

\begin{figure*}
\begin{center}
\begin{minipage}{13cm}
\centering
\begin{verbatim}

ego = new SimplexCarlaCar at start,
    with model_path globalParameters.controller_path,
    with blueprint EGO_MODEL,
    with behavior EgoBehavior(EGO_SPEED), # EgoBehavior(EGO_SPEED),
    with logger CarlaLogger(f"{globalParameters.simulation_folder}")


for i in range(5):
    random_spot = new OrientedPoint following roadDirection from start for 
            Range(DIST_PEDESTRIAN-1, DIST_PEDESTRIAN+50)

    side_offset = 8 # Range(1,8)  # Distance from road center
    side_direction = Discrete({1: 0.5, -1: 0.5})  # Left or right side randomly

    p = new Pedestrian right of random_spot by side_direction * side_offset,
        with heading side_direction * 90 deg relative to random_spot.heading,
        with regionContainedIn None,
        with behavior CrossRoadBehavior(PEDESTRIAN_MIN_SPEED, THRESHOLD)

    PEDESTRIANS += [p]

for i in range(5):
    random_spot = new OrientedPoint following roadDirection from start for 
            Range(DIST_CAR1-1, DIST_CAR1+50)
    side_movement = Discrete({0: 0.5, 3: 0.5})  # Left lane or same lane

    start_spot = new OrientedPoint left of random_spot by side_movement

    car1 = new Car following roadDirection from start_spot for Range(-1, 1),
        with behavior KeepDistance(CAR1_SPEED)

    CARS += [car1]



\end{verbatim}
\end{minipage}
\end{center}
\caption{Fragment of \textsc{Scenic} program for Use Case 2: Collision Avoidance.}
\label{scenicProgram2}
\end{figure*}

\end{document}